\PassOptionsToPackage{table}{xcolor}
\documentclass[manuscript,screen]{acmart}

\usepackage{tikz}
\usepackage{soul}
\usepackage{pgfplots}
\usetikzlibrary{pgfplots.groupplots}
\usepackage{caption}
\usepackage{subcaption}
\usepgfplotslibrary{polar}
\usepackage{multirow}
\usepackage[table]{xcolor}
\usepackage{tcolorbox}
\usepackage{svg}
\usepackage{amsmath}
\usepackage{ragged2e}
\usepackage{enumitem}
\usepackage{graphicx}
\usepackage{adjustbox}
\usepackage{booktabs}
\usepackage{array}
\usepackage{makecell}
\pgfplotsset{compat=1.18}
\definecolor{color1}{RGB}{69, 123, 157}     
\definecolor{color2}{RGB}{229, 89, 52}      
\definecolor{color3}{RGB}{144, 190, 109}    
\definecolor{color4}{RGB}{255, 159, 28}     
\definecolor{color5}{RGB}{148,103,189}
\definecolor{lightgray}{RGB}{180, 180, 180} 
\newtcolorbox{promptbox}[1]{
colback=gray!5!white,colframe=gray!70!black,fonttitle=\bfseries\sffamily,title=#1,width=0.95\textwidth,center,before={\vspace{0.1in}}}
\newtcolorbox{promptbox1}[1]{
colback=gray!5!white,colframe=gray!70!black,fonttitle=\bfseries\sffamily,title=#1,width=0.95\textwidth,center,before={\vspace{0.1in}}}

\newtcolorbox{answerbox}{
  colback=gray!15,
  colframe=black,
  arc=2mm,
  boxrule=1pt,
  left=10pt,
  right=10pt,
  top=10pt,
  bottom=10pt,
  width=0.98\textwidth,
  center,
  fontupper=\normalsize
}

\AtBeginDocument{%
  }

\setcopyright{acmlicensed}
\copyrightyear{2018}
\acmYear{2018}
\acmDOI{XXXXXXX.XXXXXXX}
\acmConference[Conference acronym 'XX]{Make sure to enter the correct
  conference title from your rights confirmation email}{June 03--05,
  2018}{Woodstock, NY}
\acmISBN{978-1-4503-XXXX-X/18/06}

\begin{document}

\title{Does Knowledge Distillation Matter for Large Language Model based Bundle Generation?}
\author{Kaidong Feng}
\email{fengkaidong@stumail.ysu.edu.cn}
\affiliation{%
  \institution{Yanshan University}
  \city{Qinhuangdao}
  \country{China}
}

\author{Zhu Sun}
\authornote{Corresponding authors}
\email{sunzhuntu@gmail.com}
\affiliation{%
  \institution{Singapore University of Technology and Design}
  \city{Singapore}
  \country{Singapore}
}

\author{Jie Yang}
\email{j.yang-3@tudelft.nl}
\affiliation{%
  \institution{Delft University of Technology}
  \city{Delft}
  \country{the Netherlands}
  }

\author{Hui Fang}
\email{fang.hui@mail.shufe.edu.cn}
\affiliation{%
  \institution{Shanghai University of Finance and Economics}
  \city{Shanghai}
  \country{China}
}

\author{Xinghua Qu}
\email{quxinghua17@gmail.com}
\affiliation{%
\institution{Bytedance Seed}
  \city{Singapore}
  \country{Singapore}
}

\author{Wenyuan Liu}
\authornotemark[1]
\email{wyliu@ysu.edu.cn}
\affiliation{%
  \institution{Yanshan University}
  \city{Qinhuangdao}
  \country{China}
}
\begin{abstract}
Large language models (LLMs) have been extensively applied in various recommendation scenarios, including bundle generation, thanks to their exceptional reasoning capabilities and comprehensive knowledge. However, exploiting large-scale LLMs for bundle generation introduces significant efficiency challenges — primarily high computational costs during fine-tuning and inference due to their massive parameterization. Knowledge distillation (KD) offers a promising solution by transferring expertise from large teacher models to more compact student models. This study systematically investigates KD approaches for bundle generation with the goal of minimizing computational demands while preserving performance.
Specifically, we explore three critical research questions: (1) how does the \textbf{\textit{format of distilled knowledge}} impact bundle generation performance? (2) to what extent does the \textbf{\textit{quantity of distilled knowledge}} influence the performance? and (3) how do different \textbf{\textit{ways of utilizing the distilled knowledge}} affect the performance? To support this investigation, we propose a comprehensive KD framework that (i) progressively extracts knowledge from raw data in increasingly complex forms, i.e., frequent patterns $\rightarrow$ formalized rules $\rightarrow$ deep thoughts; (ii) captures varying quantities of distilled knowledge through different sampling strategies, multi-domain accumulation, and multi-format aggregation; and (iii) exploits complementary LLM adaptation techniques — in-context learning, supervised fine-tuning and their combination — to leverage the distilled knowledge for domain-specific adaptation and enhanced efficiency in small student models.
Through extensive experiments on multiple real-world datasets, we provide valuable insights into how knowledge format, quantity, and utilization methods collectively shape the performance of LLM-based bundle generation, which exhibits the significant potential of KD for more efficient yet effective LLM-based bundle generation.
\end{abstract}




\maketitle

\section{Introduction}
\begin{figure}[!htbp]
    \centering
    \includegraphics[width=0.5\linewidth]{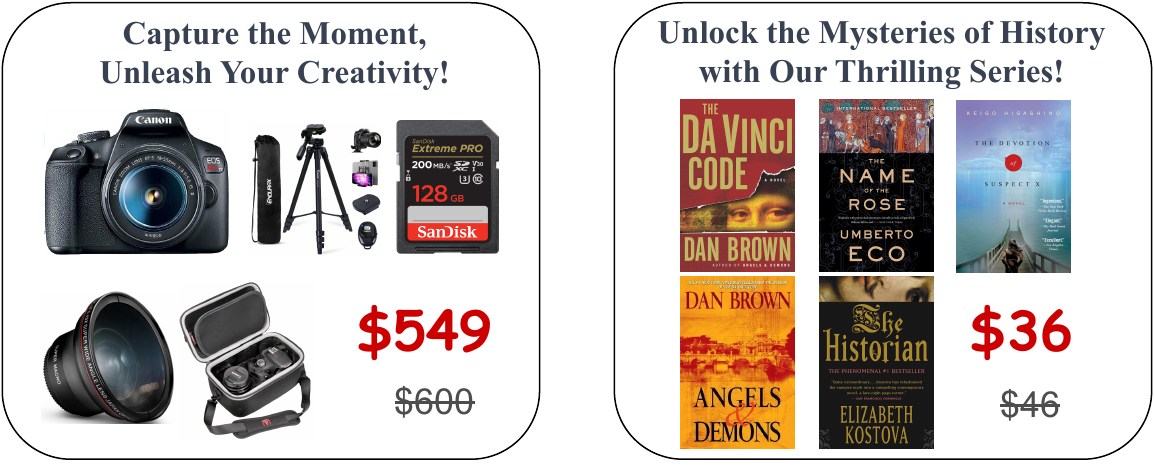}
    \caption{Example bundles for (1) a camera and its accessories; and (2) mystery, thriller, and historical fiction~\cite{sun2024adaptive}.}
    \label{fig:bundle-example}
    \vspace{-0.05in}
\end{figure}

Product bundling represents a cornerstone marketing strategy that combines multiple products or services into a single package, typically offered at a discounted price~\cite{sar2016beyond,xie2010breaking}, as illustrated in Figure~\ref{fig:bundle-example}. This strategy has demonstrated remarkable effectiveness across diverse domains like telecommunications, retail, and e-commerce~\cite{dragone2018no, sun2022revisiting}. It creates a dual benefit ecosystem: consumers gain enhanced value and convenience, while businesses experience increased sales volume and improved transaction efficiency~\cite{harris2006consumer}. Given these compelling advantages, product bundling has garnered significant research attention, particularly in the area of bundle recommendation — a field focused on suggesting curated item sets (i.e., bundles) to users based on their preferences, assuming the pre-existence of bundles. In particular, they treat either co-purchased products ~\cite{liu2017modeling, he2022bundle} or user-generated lists~\cite{he2019hierarchical,chang2020bundle,he2020consistency} as bundles, or directly use pre-defined bundles by retailers~\cite{chen2019pog}. However, the quality of bundles used in these methods is often suboptimal due to limitations such as noise in co-purchased data, lack of diversity in user-generated lists, and the high cost and expertise required for creating pre-defined bundles. 

These limitations have led to the emergence of the bundle generation task, which is specifically dedicated to constructing high-quality bundles. Early studies utilized constraint-based methods, specifying conditions like budget limits or maximizing metrics such as savings, customer adoption, or expected revenue~\cite{xie2010breaking,xie2014generating,garfinkel2006design,yang2012bundle,beladev2016recommender,zhu2014bundle}. Subsequent methods leverage neural network models (e.g., attention mechanism) to learn relationships (e.g., alternative or complementary) among items to form bundles~\cite{bai2019personalized,chang2021bundle,wei2022towards}. Nevertheless, significant challenges persist as existing methods demonstrate critical limitations in generating bundles with necessary diversity and flexible sizing while simultaneously aligning them with the nuanced spectrum of consumer intentions and contextual needs. Recent advancements in large language models (LLMs) have demonstrated transformative capabilities within recommender systems (RSs), excelling at preference understanding, recommendation reasoning, and personalized content generation ~\cite{wang2023generative, wang2023enhancing}. These approaches fundamentally reconceptualize recommendation tasks as natural language problems, enabling LLMs to produce sophisticated recommendations in textual formats. This paradigm shift has naturally extended to bundle generation~\cite{sun2024adaptive}, where LLMs' reasoning capabilities and contextual understanding offer promising solutions to longstanding challenges in creating coherent, diverse, and intent-aligned bundles.

While LLMs show exceptional performance in bundle generation, their massive parameterization creates significant challenges: prohibitive computational overhead, deployment complexities, and extended inference times resulting in high latency. These limitations become particularly critical in RSs where real-time responsiveness and frequent model updates are essential operational requirements. To mitigate these issues, \textbf{\textit{knowledge distillation}} (KD) has emerged as a prominent technique by transferring knowledge from a larger, more complex teacher model to a smaller, more efficient student model without significantly sacrificing performance. Inspired by this, our study systematically investigates KD approaches for LLM-driven bundle generation with the dual objectives of substantially reducing computational demands while maintaining generation quality. We explore three fundamental research questions:
\begin{itemize}
    \item \textbf{RQ1: How does the format of distilled knowledge affect the performance on bundle generation}? We examine whether different types of knowledge yield varying effectiveness when transferred to student LLMs.
    \item \textbf{RQ2: To what extent does the quantity of distilled knowledge affect the performance on bundle generation}? We investigate the relationship between knowledge volume and bundle generation quality to identify optimal efficiency-effectiveness trade-offs for student LLMs.
    \item \textbf{RQ3: How do different ways of utilizing the distilled knowledge affect the performance on bundle generation}? We compare various knowledge integration approaches to determine their relative effectiveness in preserving critical capabilities of student LLMs.
\end{itemize}
%

To answer these research questions, we design a comprehensive KD framework to investigate how knowledge format, quantity, and utilization methods collectively shape the performance of LLM-based bundle generation, establishing a foundation for more efficient bundle generation architectures without sacrificing quality.
First, it progressively extracts knowledge from the raw data in increasingly sophisticated forms, ranging from frequent patterns to formalized rules and deep thoughts specifically optimized for bundle generation tasks. Next, it captures varying quantities of distilled knowledge through various sampling {(i.e., random-, length-, diversity- and difficulty-based)} strategies, multi-domain accumulation and multi-format aggregation, enabling systematic evaluation of knowledge volume-performance relationships. Finally, it exploits complementary LLM adaptation techniques, i.e., in-context learning (ICL), supervised fine-tuning (SFT), and their combination, to leverage the distilled knowledge for domain-specific adaptation in lightweight, small student LLMs (i.e., Llama3.1-8b), significantly reducing computational requirements while maintaining generation quality comparable to larger language models.  

Several major findings are gained through our extensive experiments on three real-world datasets. For instance,  
regarding \textbf{RQ1}, different formats of distilled knowledge positively impact bundle generation performance by enhancing student models, with SFT
benefiting more consistently and even surpassing the teacher model in Precision and Coverage, though challenges remain in achieving comparable Recall. For \textbf{RQ2}, increasing the quantity of distilled knowledge positively impacts bundle generation performance, with ICL benefiting more from higher sampling ratios due to its reliance on a larger knowledge pool for retrieval, while SFT tends to peak at a certain sampling ratio and gains substantial improvements from aggregating knowledge across different formats and domains. For \textbf{RQ3}, bundle generation performance significantly depends on the knowledge utilization method, with combining SFT with ICL generally being most effective when knowledge types are carefully selected at each stage, though SFT alone with distilled knowledge also provides great potential. Additionally, among the three factors, the utilization method investigated in RQ3 shows the most significant impact on the student model performance, whereas the knowledge format examined in RQ1 contributes the least. 
%
%
Overall, our contributions are highlighted as follows:
\begin{itemize}
    \item We, for the first time, conduct a systematic exploration of knowledge distillation techniques for the bundle generation task, addressing the critical challenge of computational overhead in LLM-based bundle generation while maintaining generation quality.
    \item We formulate three research questions to guide our exploration. To address these questions, we propose a comprehensive knowledge distillation framework that (1) progressively extracts knowledge in increasingly sophisticated forms, (2) designs various strategies to capture varying quantities, and (3) employs various LLM adaptation techniques, to investigate how knowledge format, quantity, and utilization methods influence the performance of bundle generation. 
    \item We conduct extensive experiments across three real-world datasets, whereby we identify solutions that leverage small student models to generate bundles of comparable quality to larger language models while significantly reducing computational requirements. Additionally, our empirical exploration yields valuable insights into how different characteristics of distilled knowledge collectively shape the performance of bundle generation, offering essential guidance for future research. 
\end{itemize}

\section{Related Work}
This section briefly reviews the related literature for our work, consisting of bundle recommendation and generation, large language models (LLMs) for recommendation, and knowledge distillation (KD) in recommendation.

\subsection{Bundle Recommendation}
Early research on bundle recommendation mainly exploits \textit{conventional methods}. For instance, constrained-based methods take into account different practical constraints like cost, revenue across different scenarios such as e-commerce~\cite{zhu2014bundle} and travel package recommendation~\cite{xie2010breaking,liu2011personalized}. Data-mining approaches leverage association rules~\cite{fang2018customized} or probabilistic models~\cite{liu2017modeling} to discover bundle patterns. Preference elicitation methods focus on learning utility functions~\cite{xie2014generating,dragone2018no} to capture user preferences across various features, while factorization-based approaches like LIRE~\cite{liu2014recommending}, BBPR~\cite{pathak2017generating}, and EFM-Joint~\cite{cao2017embedding} decompose user-item and user-bundle matrices to learn underlying preferences. 
Subsequently, \textit{deep learning-based methods} show their stronger capabilities in bundle recommendation. Accordingly, different neural network architectures have been proposed. Specifically, sequence-based methods like BGN~\cite{bai2019personalized} and  ComEmb~\cite{kouki2019product} utilize LSTM to build bundles dynamically and then recommend these bundles to users. Attention-based approaches like DAM~\cite{chen2019matching}, AttList~\cite{he2019hierarchical}, CAR~\cite{he2020consistency}, and BRUCE~\cite{avny2022bruce} learn item affinity or user preference towards bundles using the attention mechanism. Graph-based methods, such as BundleNet~\cite{deng2020personalized} and BGCN~\cite{chang2020bundle}, capture complex user-item-bundle relationships through graph convolutional networks. Recent advances include contrastive learning methods (e.g., MIDGN~\cite{zhao2022multi}, HIDGN~\cite{zou2023towards}, CrossCBR~\cite{ma2022crosscbr}, BundleGT~\cite{wei2023strategy}, and MultiCBR~\cite{ma2024multicbr}) for more effective representation learning and conversational approaches (e.g., BUNT~\cite{he2022bundle}) that enable dynamic recommendation through multi-round interactions. Besides, {CoHEAT~\cite{jeon2024cold} employs curriculum learning to dynamically balance user-bundle and user-item interactions based on bundle popularity to address the cold-start problem in bundle recommendation.} Despite their effectiveness, most of these methods directly regard co-purchased products~\cite{zhu2014bundle,liu2017modeling} or user-generated lists~\cite{he2019hierarchical,chang2020bundle,he2020consistency,chen2019matching,dragone2018no,liu2014recommending} as bundles. However, many products are co-purchased with no common underlying intents, so there is no quality control to ensure bundle coherence. Besides, the user-generated lists are only available in limited domains, e.g., books and music. Although some studies utilize pre-defined bundles by retailers~\cite{ge2014cost,fang2018customized,bai2019personalized,liu2011personalized,deng2020personalized}, the data size is limited due to the high cost of collecting.

\subsection{Bundle Generation}
The demand for high-quality bundles has driven significant research in bundle generation, evolving from simple pattern mining to modern multimodal approaches. Early studies~\cite{sun2022revisiting} focus on mining co-occurrence patterns at the item category level. The field then advanced to include user preferences, marking a shift toward more personalized solutions. Specifically, POG~\cite{chen2019pog} employs the Transformer architecture to capture item compatibility relationships based on the image and textual information of items and learn from user interaction history to generate personalized outfits that match user preferences. BYOB~\cite{deng2021build} treats bundle generation as a combinatorial optimization problem and adopt reinforcement learning techniques to improve personalization. Recent approaches have increasingly leveraged multimodal information and user feedback. For instance, CLHE~\cite{ma2024leveraging} uses self-attention mechanisms to fuse multiple data modalities, and Conna~\cite{wei2022towards} incorporates a contrastive non-autoregressive decoding strategy to improve both creative quality and generation efficiency. Cross-item relationships have been explored by CIRP~\cite{ma2024cirp}, which integrates item semantics and relations into multimodal representations. DiFashion~\cite{xu2024diffusion} utilizes the diffusion model to synthesize fashion item images that compose a visually compatible outfit. {By incorporating a trainable fusion module, BundleLLM~\cite{liu2025harnessing} aligns multimodal features in the LLM semantic space, allowing it to process multiple data formats and complete items in the partial bundles.}
However, these methods still face critical limitations: (1) most methods can only generate fixed-size bundles, lacking the flexibility to create bundles of varying sizes based on different scenarios, and (2) some approaches require a partial bundle as context, limiting their practical applications. A recent approach, AICL~\cite{sun2024adaptive}, leverages LLMs with in-context learning techniques to generate bundles of dynamic lengths while inferring user intent. Although this method addresses the aforementioned limitations, the utilization of LLMs brings new challenges related to computational resource demands and response latency.

\subsection{LLMs for Recommendation}
Recently, LLMs have been widely applied into recommender systems (RSs) thanks to their superior reasoning capabilities and extensive knowledge~\cite{zhang2023chatgpt, wu2023survey, harte2023leveraging,sun2024large}, which can be divided into two paradigms: \textit{prompting-based} and \textit{tuning-based methods}. 
In particular, prompting-based methods keep the original LLM parameters while designing effective demonstrations through in-context learning (ICL) to improve performance on specific tasks. For example, KP4SR~\cite{zhai2023knowledge} transforms the structured knowledge graph into knowledge prompts for sequential recommendation. Other methods~\cite{dai2023uncovering,he2023large} adopt LLMs for different recommendation tasks (e.g., ranking and conversation) by reformulating them into prompt formats. In contrast, the tuning-based methods utilize the parameter-efficient fine-tuning (PEFT) to update a small number of parameters of LLMs, injecting collaborative signals or other side information. These methods transform traditional user-item interaction data into textual prompts, then fine-tune LLMs to enhance performance across various downstream recommendation tasks, such as TallRec~\cite{bao2023tallrec}, GLRec\cite{wu2024exploring}, and Rella~\cite{lin2024rella}. Although they have demonstrated the potential of LLMs in various recommendation scenarios, the integration of LLMs into bundle generation has been under-explored.
To the best of our knowledge, AICL~\cite{sun2024adaptive} is among the few methods that have applied LLMs to bundle generation, which utilizes retrieval-augmented generation and chain-of-thought techniques to generate effective demonstrations, enabling user intent inference and personalized bundle generation. However, this method faces challenges in computational efficiency due to its reliance on resource-intensive LLMs.

\subsection{Knowledge Distillation in Recommendation}
Knowledge distillation (KD) is first proposed in~\cite{hinton2015distilling}, which trains a compact student model to mimic a larger teacher model's soft probability outputs on MNIST digit classification tasks. This approach aims to transfer knowledge from an ensemble or complex model into a more lightweight and efficient model. Following its success in computer vision, KD has been widely adopted into RSs in two distinct ways: \textit{explicit} and \textit{implicit} knowledge transfer. 
The former methods take the final outputs of the teacher model as supervisory labels, training student models to produce similar results. For example, SLIM~\cite{wang2024SLIM} and RDRec~\cite{wang2024rdrec} distill the capability of generating rationales behind users' behaviors from LLMs to smaller student models using chain of thought~\cite{wang2023self}. Similarly, DLLM2Rec~\cite{cui2024DLLM2Rec} transfers ranking and collaborative knowledge from  LLM-based teacher models to conventional sequential-based student recommendation models. 
Conversely, the latter methods enhance the feature representations of student models by aligning outputs between teacher and student models at hidden layers or output layers. For instance, LaMP~\cite{salemi2024lamp} minimizes the KL-divergence between the probability distribution of student model and a target distribution computed from the teacher model for personalized text generation. PRM-KD~\cite{wen2024prmkd} distills knowledge from multiple teacher models to a student model by minimizing the KL-divergence between probability distributions in scoring and ranking items, focusing on in-batch negative samples while dynamically weighing each teacher's contribution based on confidence and mutual agreement. Additionally, some methods like NewsBERT~\cite{wu2021newsbert}, Tiny-Newsrec~\cite{yu2021tiny}, and SSI~\cite{yuan2021improving}, aim to align with not only the teacher models' probability distribution but also with the intermediate layers' representations. Inspired by the remarkable success of knowledge distillation (KD) methods in various recommendation tasks, our study explores how to leverage KD to achieve both effective and efficient bundle generation using LLMs\footnote{We focus solely on explicit knowledge distillation because most state-of-the-art LLMs (e.g., ChatGPT), are accessible only through APIs, which do not provide access to logits or intermediate representations. Additionally, open-source LLMs require substantial computational resources to achieve comparable performance. Therefore, we do not consider implicit knowledge distillation in our current work.}. 


\section{The Proposed Knowledge Distillation Framework}
\begin{figure}[t]
  \centering
  \includegraphics[width=0.85\linewidth]{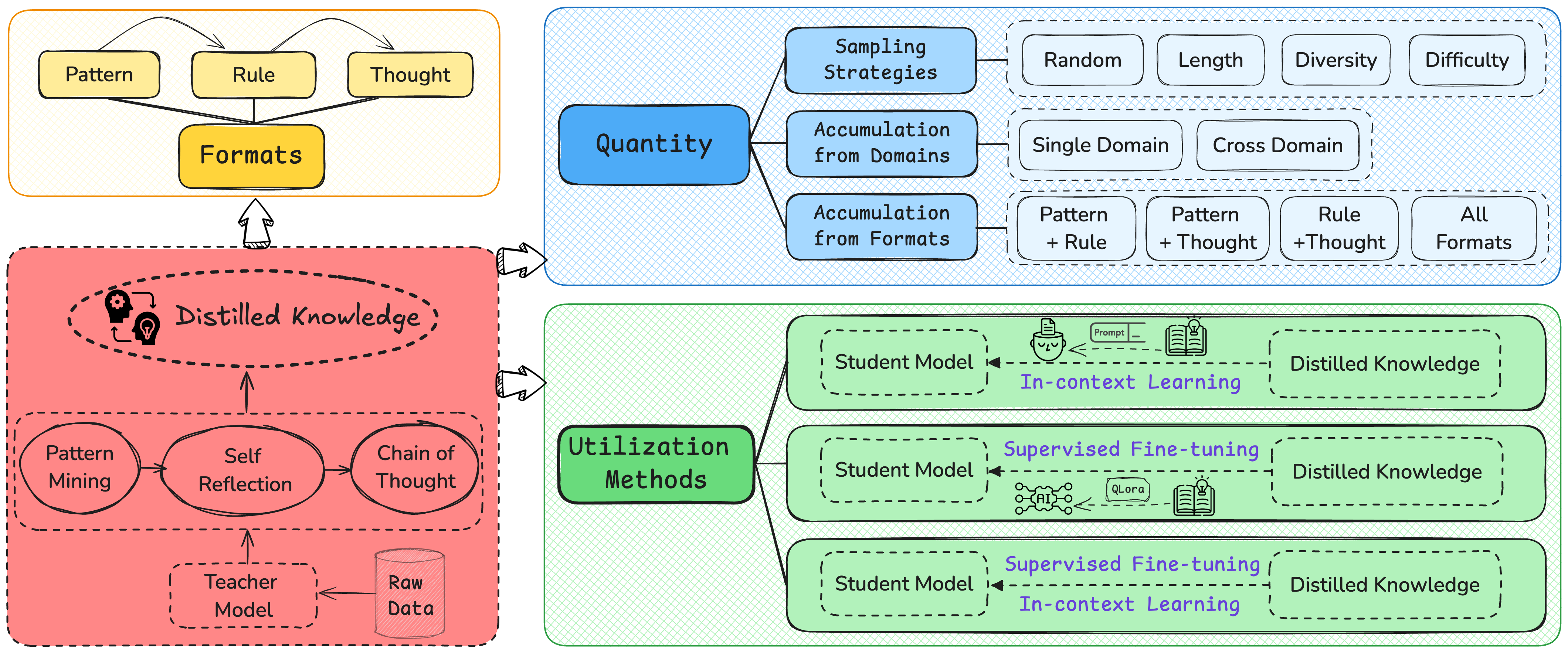}
  \caption{The overview of our proposed knowledge distillation framework. 
  }
  \label{fig:framework}
  \vspace{-0.05in}
\end{figure}

As illustrated in Figure~\ref{fig:framework}, we propose a comprehensive knowledge distillation (KD) framework to systematically investigate the impacts of the format, quantity, and utilization methods of distilled knowledge (corresponding to RQ1-RQ3) in optimizing lightweight LLMs for efficient yet effective bundle generation.

\subsection{Formats of Distilled Knowledge (RQ1)}
To investigate the impact of different types of knowledge on bundle generation, we first introduce a progressive knowledge distillation module that systematically extracts knowledge from the raw data in increasingly sophisticated forms, evolving from frequent patterns to formalized rules and deep thoughts. 

\subsubsection{Raw Data}\label{sec:raw-data}
We first introduce the datasets (i.e., raw data) from which different types of knowledge are distilled. In particular, we adopt three public bundle datasets released in the SIGIR 2022 resource paper~\cite{sun2022revisiting,sun2024revisiting}. The reasons behind lie two-fold: (1) they contain high-quality bundles and the corresponding user intents annotated through a carefully designed crowd-sourcing task based on the Amazon datasets~\cite{he2016ups} across three domains, namely Electronic, Clothing, and Food; (2) compared to other public bundle datasets, such as Steam, Netease, Youshu~\cite{avny2022bruce}, Goodreads~\cite{he2020consistency}, and iFashion~\cite{ren2023distillation}, our selected datasets provide rich side information, including user sessions (where the bundles are identified), well-labeled user intents, item titles, images, and categories, which are more suitable for our study. The statistics of the datasets are summarized in Table~\ref{tab:datasets}.   
Given the datasets, \textbf{the bundle generation task is formulated as identifying potential bundles within each session}.


\begin{table}[t]
\centering
\caption{The statistics of the three bundle datasets.}
\label{tab:datasets}
\vspace{-0.15in}
\begin{tabular}{l|cccccc}
\toprule
 & \#Users & \#Items & \#Sessions & \#Bundles & \#Intents & Average Bundle Size \\
\midrule
Electronic & 888 & 3499 & 1145 & 1750 & 1422  & 3.52 \\
Clothing & 965 & 4487 & 1181 & 1910 & 1466  & 3.31 \\
Food & 879 & 3767 & 1161 & 1784 & 1156  & 3.58 \\
\bottomrule
\end{tabular}
\vspace{-0.1in}
\end{table}

\subsubsection{Frequent Patterns} 
Frequent patterns represent the first level of our distilled knowledge, marking a significant advancement in abstraction from raw data. While raw data merely provides isolated information about individual items within bundles, frequent patterns uncover meaningful relationships that link these items within coherent bundles, transforming our understanding from merely "what items exist" to "what items belong together". Inspired by prior works~\cite{sun2022revisiting}, we attempt to identify frequent patterns at the item category level across different bundles by employing the Apriori algorithm~\cite{agrawal1994fast}. 
Note that we do not use teacher LLMs to distill the frequent patterns due to the content length constraint. 
Specifically, we transform all bundles (e.g. $b_1=\{i_1, i_2, i_3\}$) into their categorical representations (e.g. $b_1=\{c_1, c_2, c_3\}$) and apply Apriori algorithm to mine frequent patterns (i.e., frequently co-occurred categories) across different bundles (e.g. $p_1 = \{c_1, c_2\}$ or $p_2 = \{c_1, c_2, c_3\}$), where $b, i, c, p$ denote a bundle, an item, an item category and a pattern, respectively.
These extracted frequent patterns thus serve as pre-processed insights and by explicitly providing such patterns to the student LLM, we enable it to recognize that specific category combinations (complementary or alternative items) consistently appear in high-quality bundles. In summary, the advantage of leveraging pattern knowledge is twofold: it reduces the cognitive load on LLMs by providing distilled category-level heuristics, and it acts as a structural guide for bundle generation. When prompted or fine-tuned with pattern knowledge, LLMs can more effectively generate coherent bundles by first selecting appropriate category combinations and then populating those categories with contextually relevant items in user sessions, resulting in bundles that better mirror real-world user preferences and behaviors. Table~\ref{tab:frequent-patterns} below demonstrates examples of frequent patterns mined from the three domains. 

\begin{table}[!htbp]
\centering
\caption{Examples of frequent patterns mined on the three domains.
}
\label{tab:frequent-patterns}
\vspace{-0.1in}
\renewcommand{\arraystretch}{1.2}
\begin{tabular}{@{}l|p{4.5cm}|p{3.5cm}|p{4cm}@{}}
\toprule
Domains & Electronic & Clothing & Food \\
\midrule
\multirow{2}{*}{\makecell[l]{Frequent \\ Patterns}} &
[Camera, Camera Batteries, Camera Lenses, Micro SD Cards] &
[Sandals, Cover-Ups, Handbags, Hats \& Caps] &
[Baking Cups, Crackers, Peanut Butter, Toaster Pastries] \\
\specialrule{.05em}{.05em}{.05em}
\makecell[l]{Corresponding \\ Bundles} &
\adjustbox{valign=m}{\includegraphics[width=3.5cm]{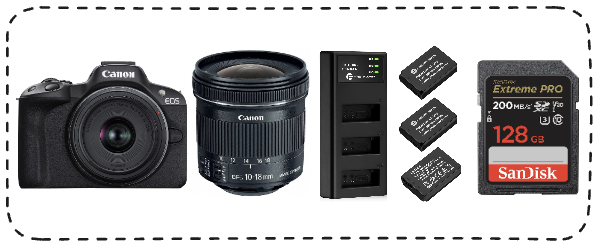}} &
\adjustbox{valign=m}{\includegraphics[width=3.5cm]{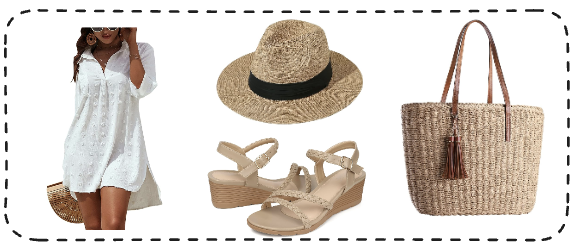}} &
\adjustbox{valign=m}{\includegraphics[width=3.5cm]{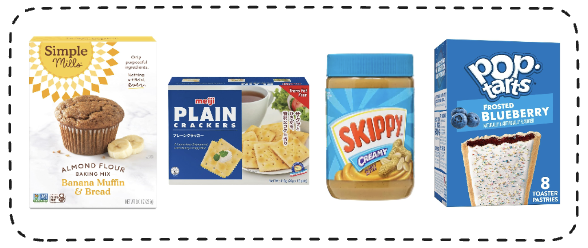}} \\
\bottomrule
\end{tabular}
\vspace{-0.1in}
\end{table}


\subsubsection{Formalized Rules}
Rules represent the second level of our distilled knowledge, transforming frequent patterns into explicit and actionable principles. While pattern knowledge identifies relationships between item categories, rule knowledge establishes clear conditions and guidelines for effective bundle generation. This shift advances understanding from simply recognizing “what items belong together” to reasoning about “why and when items should be bundled.” 

To distill rule knowledge, we employ the self-reflection method~\cite{madaan2023self} with teacher LLMs. For each session, we first design a prompt to ask LLMs to generate initial bundles as - \textcolor{black!70}{\textit{A bundle can be a set of alternative or complementary products that are purchased with a certain intent. Given the list of products with their descriptions: \{product X: category, title\}, identify bundles where: - Each bundle must contain at least 2 products; - Products must serve a common user intent; - Products must be either complementary (work together) or alternative (substitutes). Answer format in JSON: \{`bundle number':[`product number']\}. Please do not provide any explanations for the results}}. Then, we provide the ground truth bundles and ask LLMs to compare their predictions with the correct bundles, analyzing the reasons behind any discrepancies using the prompt: \textcolor{black!70}{\textit{Compare the correct bundles and your answers: Correct bundles: \{correct bundles\}. Your answers: \{detected bundles\}. Identify which detected bundles in your answers are incorrect and explain why they are incorrect based on the product categories and descriptions. Answer format in JSON: \{`incorrect bundle number':[`reason for incorrect detection']\}}}. Next, we ask LLMs to review the entire process, identifying the underlying causes of incorrect bundle predictions with the prompt: \textcolor{black!70}{\textit{Review your bundle detection process considering the following: 1. User Intent Analysis: How well did you identify the primary intent? 2. Product Relationships: The products within a bundle should have a certain relationship (e.g., alternatively or complementarily). 3. Bundle Logic: Each bundle should contain at least 2 products; Practical usage scenarios that reflect certain topics or user intents. Answer format in JSON: \{`issue number':[`specific aspect', `detailed reasoning with example']\}}}. Finally, based on this analysis, we prompt LLMs to summarize explicit rules for correct bundle generation: \textcolor{black!70}{\textit{Based on your analysis of correct and incorrect bundles, formulate the rules that should be followed to improve the accuracy of bundle detection. Each rule should: 1. Define specific criteria for grouping (e.g., complementary relationships or alternative options). 2. Avoid common mistakes (e.g., unrelated products in a bundle). 3. Align with purchase intent. Answer format in JSON: \{`rule number':[`rule description']\}}}. Unlike patterns that merely reflect statistical co-occurrence, rules capture causal relationships and conditional logic, which explain why certain item combinations succeed while others fail, thus enabling LLMs to make more natural bundling decisions for more principled and effective bundle generation. Table~\ref{tab:rules} showcases examples of rules derived from the three domains. 

\begin{table}[!htbp]
\centering
\caption{Examples of formalized rules derived from the three domains.
}
\label{tab:rules}
\vspace{-0.15in}
\begin{tabular}{l|l}
\toprule
Domain &Rules \\\midrule
Electronic & 1. Group products together if they are complementary and serve a common user intent, such as \\ 
&protective cases and batteries for electronic devices. \\ &2. Aim to create bundles with at least 2 products that reflect practical usage scenarios aligned with \\
&specific topics or user intents.\\
\specialrule{.05em}{.05em}{.05em}
Clothing &  1. Avoid combining products in a bundle if they are unrelated in terms of user intent or \\
&practical usage scenarios (e.g., Belts and Earmuffs)\\
&2. Group products based on shared use or user intent, such as creating a coordinated jewelry \\
&accessories set for a specific occasion.\\
\specialrule{.05em}{.05em}{.05em}
Food & 1. Avoid grouping products that are unrelated or do not align with a common user intent, such as \\
&mixing nutrition bars with mints.\\
&2. Ensure that products in a bundle belong to related or overlapping categories to avoid mismatched \\
&pairings like coffee capsules with coffee pod holders.\\
\bottomrule
\end{tabular}
\vspace{-0.1in}
\end{table}

\subsubsection{Deep Thoughts} Thoughts are the highest level of our distilled knowledge, which introduce a more flexible, context-sensitive approach to bundle generation. While rule knowledge provides guidelines and principles for bundle generation, thought knowledge enables dynamic reasoning that adapts to specific scenarios. This transition shifts the focus from “why items should be bundled” to “how to optimize bundle generation for unique situations”.

To distill thought knowledge, we utilize chain-of-thought (CoT)~\cite{wang2023self} reasoning with teacher LLMs. For each session, given its item information (e.g., title, category), all possible bundles, and the corresponding user intents, we prompt LLMs to explain the reasons why a particular bundle is formed, following a complete reasoning path — from meta-information (categories) to user intent and bundle composition. 
\begin{itemize}[leftmargin=*]
    \item[-] \textcolor{black!70}{\textit{You are provided with: 1. A product list containing products with their categories: \{`product id':\{`title': `$\$title$', `category': `$\$category$'\}\} 2. A list of bundles containing groups of products commonly purchased together with the user intents, detected from the product list: \{`bundle $id$':\{`group': [`\$product $id$'], `intent': `$\$intent$'\}\}.}}
    \item[-] \textcolor{black!70}{\textit{Your Task: The products in a bundle are commonly purchased together based on a specific user intent, and all the bundles are detected from the product list. Generate natural language insights for each bundle explaining why certain product categories are bundled together based on user intent.}}
    \item[-] \textcolor{black!70}{\textit{Output Format: \{`bundle1': `Customers buying [category a] and [category b] are typically looking to [intent]', `bundle2': `The combination of [category x] and [category y] suggests [intent]'.\}}}
    \item[-] \textcolor{black!70}{\textit{Example Output: \{`bundle1': `Gaming enthusiasts purchase GPU, CPU, and Motherboard together for building high-performance gaming setups.', `bundle2': `Customers combining Office Chairs and Monitors are focused on creating an ergonomic and efficient workspace.'\}}}
    \item[-] \textcolor{black!70}{\textit{Requirements: 1. Insightful Explanations: Derive meaningful connections between product categories and user intents. 2. Clarity and Natural Tone: Write easy-to-understand, conversational explanations. 3. Category-Intent Pairing: Each explanation must include references to product categories and intents. 4. Deliver only JSON output with explanations. 5. Avoid unnecessary details or supplementary text.}}
\end{itemize}

This step-by-step process articulates the complex considerations behind effective bundling, capturing subtle nuances often missed by simpler knowledge types. Thought knowledge equips LLMs to move beyond fixed patterns and rules, integrating contextual intelligence into bundle generation. It enables student LLMs to reason dynamically, evaluate multiple factors concurrently, and adapt their strategies based on specific user needs and contexts, leading to more intuitive and effective bundling decisions. Examples of derived thoughts from the three domains are shown in Table \ref{tab:thoughts}.

\begin{table}[!htbp]
\centering
\caption{Examples of deep thoughts derived on the three domains.
}
\label{tab:thoughts}
\vspace{-0.15in}
\begin{tabular}{l|l}
\toprule
Domain &Thoughts \\\midrule
Electronic & 1. Customers purchasing \textit{Screen Protectors} and \textit{Folio Cases} together are likely looking to protect and \\& accessorize their MacBook Pro 13.3 with Retina Display.\\
&2. Customers buying \textit{Film, Camera Batteries}, and \textit{Skylight \& UV Filters} are typically looking to  \\& enhance their photography equipment with essential accessories for better image quality. \\
\specialrule{.05em}{.05em}{.05em}
Clothing &  1. Customers purchasing \textit{Hats} and \textit{Winter Accessories} are likely preparing for cold weather and\\& looking to stay warm and stylish.\\
&2. The combination of \textit{Dresses, Pumps}, and \textit{Costumes} suggests customers are preparing for various \\ &special events or themed parties.\\
\specialrule{.05em}{.05em}{.05em}
Food & 1. Customers buying \textit{Cakes} and \textit{Gluten Free} products are likely looking for baking options.\\
&2. Customers buying \textit{Fruit Snacks} and \textit{Oatmeal} together are likely looking for convenient and healthy \\&snack options for on-the-go consumption.\\
\bottomrule
\end{tabular}
\vspace{-0.1in}
\end{table}

\input{chart/knowledge_scale}

\subsection{Quantity of Distilled Knowledge (RQ2)}\label{sec:quantity-of-knowledge}  
To explore how the quantity of distilled knowledge impacts bundle generation performance, we design four distinct sampling strategies (i.e., random-, length-, diversity-, and difficulty-based) to select specific portions of the entire raw data. These strategies aim to capture varying knowledge quantities while ensuring representative data coverage for effective and efficient KD. Note that this process assumes that varying data portions will yield different amounts of distilled knowledge.

\subsubsection{Different Sampling Strategies}
\textit{Random-based strategy} randomly samples a portion of raw data to distill different formats of knowledge. 
\textit{Length-based strategy} first divides the entire raw data into three groups based on session length, with defined ranges of [2-4], [5-7], and [8-10], and then samples a portion of raw data from each group. The rationale behind this is that sessions of varying lengths may capture different levels of user interaction complexity and encompass diverse numbers of bundles. 
\textit{Diversity-based strategy} first divides the entire raw data into three groups (i.e., low, medium, and high) based on session diversity levels, which are measured by the ratio of the number of categories to the number of items within each session. It then samples a portion of raw data from each group.
\textit{Difficulty-based strategy} first divides the entire raw data into three groups (i.e., easy, medium, and hard) based on the difficulty levels for bundle generation. Specifically, we leverage teacher LLMs with the zero-shot setting to perform the bundle generation task (i.e., identify bundles from each session) and define the difficulty based on the accuracy of the identified bundles. Then, it samples a portion of raw data from each group.  
To investigate the impact of varying amounts of distilled knowledge, we apply these sampling strategies with different sample ratios, specifically in the range of \{10\%, 30\%, 50\%, 70\%\}.

\smallskip\noindent\textbf{Assumption Verification}. As emphasized, our approach is based on the assumption that varying the data portions will result in different amounts of distilled knowledge. For verification, we apply the four sampling strategies with sample ratios in the range of \{10\%, 30\%, 50\%, 70\%\} on the raw data. Then, we distill knowledge from the sampled data, and calculate the amounts of the distilled knowledge. Figure~\ref{fig:knowledge_scale} shows the results, where the x-axis represents the sampling ratio and the y-axis means the amount of distilled knowledge. Specifically, for pattern knowledge, we merge the same patterns (e.g., $[c_1, c_2]$ and $[c_2, c_1]$) to eliminate duplicates. For rule and thought knowledge expressed as natural language, we use a pre-trained model (i.e., BERT) to compute the semantic similarity and filter out redundant rules or thoughts based on a similarity threshold of 0.8.
The results show that as the sample ratios increase, the amount of distilled knowledge also increases across the four strategies and three domains, validating our assumption.  


\subsubsection{Accumulation of Knowledge from Different Domains and Formats}
In addition to employing different sampling strategies to vary the quantity of distilled knowledge, we explore two additional approaches. The first method focuses on accumulating a single type of knowledge from multiple domains, enabling a comparison between using a specific type from a single domain (e.g., rules from Electronic) and combining such type of knowledge from multiple domains (e.g., rules from Electronic + Clothing + Food).
The second method involves accumulating diverse types of knowledge, allowing us to compare the performance of using a single knowledge type (e.g., frequent patterns) with that of multiple types (e.g., frequent patterns + formalized rules + deep thoughts).  In summary, the first method incorporates cross-domain knowledge to enhance the student LLM's generalizability, whereas the second method integrates diverse and rich knowledge from a single domain into the student LLM, enabling it to acquire more comprehensive and in-depth domain-specific knowledge.

\subsection{Utilization Methods of Distilled Knowledge (RQ3)}\label{sec:utilizing-methods}
Given the distilled knowledge in various formats and quantities, we further explore 
the impacts of different methods for utilizing distilled knowledge on bundle generation performance. To this end, we consider two complementary LLM adaptation techniques: in-context learning (ICL) and supervised fine-tuning (SFT). Using these techniques, we examine three distinct utilization methods on the student LLMs: (1) incorporating distilled knowledge into ICL for inference, (2) integrating distilled knowledge through supervised fine-tuning for training, and (3) injecting distilled knowledge into both SFT for training and ICL for inference simultaneously.

\subsubsection{In-Context Learning (ICL)}\label{sec:icl-prompt}
We propose various knowledge retrieval methods based on the features of the target session and incorporate the retrieved knowledge into the prompt as context to enhance the student LLM's task comprehension. 
For pattern knowledge, we employ a fine-grained category matching strategy. Given a session, we retrieve all patterns that form a subset of the categories involved in the target session. The retrieved patterns are then integrated into the prompt as context.
For rules and thoughts, we first generate semantic embeddings of product titles within each session using a pre-trained language model (BERT). Next, we retrieve the session most similar to the target session based on the cosine similarity of their semantic embeddings. The rules and thoughts associated with the retrieved session are incorporated into the prompt as context.
The final prompt combines the target session, its product information, and the retrieved knowledge as context, given below:
\begin{itemize}[leftmargin=*]
    \item[-] \textcolor{black!70}{You will be given a list of products, each with a category and description. You will also receive a single piece of overarching guidance in the variable $KNOWLEDGE$. }
    \item[-] \textcolor{black!70}{Your task is to: Analyze the products and identify relationships between them, referencing $KNOWLEDGE$ to determine their connections. Group the products into bundles based on these criteria: 1. Each bundle must contain at least 2 products; 2. Products in a bundle should be either alternative options for the same purpose or complementary items typically bought together; 3. The products should fit a specific customer intent or use case indicated by $KNOWLEDGE$. Present the bundles in JSON format as follows: $\{$`bundle1': [`product1', `product2', \dots], `bundle2': [`product3', `product4', \dots], \dots$\}$}
    \item[-] \textcolor{black!70}{Important notes: 1. Only provide the JSON output; 2. Do not include any explanations or additional text; 3. Use `bundle1', `bundle2', etc. as keys in the JSON; 4. Use the product IDs as values in the arrays; 5. Leverage the guidance from $KNOWLEDGE$ to remain consistent with the bundling logic.}
\end{itemize}

\subsubsection{Supervised Fine-Tuning (SFT)} 
This method enables student LLMs to acquire task-specific knowledge, thereby mitigating the hallucination problem.
To implement this method, we construct training data using the same textual prompt in Section~\ref{sec:icl-prompt} (i.e., including session information and retrieved knowledge) as input, and the ground truth bundles from the session as output. To further enhance the robustness of student LLMs, we introduce a {permutation augmentation} strategy during training data construction.
Since a session may involve multiple bundles, this strategy perturbs the bundle order, allowing the student LLM to learn order-invariant properties among bundles. Specifically, for each training sample, we generate all possible bundle permutations to form an equivalent sample set. For example, given a session $s_1$, if the ground truth bundles (label) are $\{b_1:[i_1,i_2], b_2:[i_3,i_4]\}$, applying our permutation strategy will also include $\{b_2:[i_3,i_4], b_1:[i_1,i_2]\}$.

\section{Experiments, Results and Analysis}
\subsection{Experimental Settings}
 
\textbf{Datasets}.
We conduct experiments with three public bundle datasets, as introduced in Section~\ref{sec:raw-data}. Specifically, we split all session data into training, validation, and test sets with a ratio of 7:1:2.  

\smallskip\noindent\textbf{Teacher \& Student  Models}. 
We use GPT-3.5-turbo as the teacher LLM and Llama3.1-8B as the student LLM. Based on this setup, we construct \textbf{Llama3.1-ICL} and \textbf{Llama3.1-SFT} by integrating distilled knowledge from the teacher model through in-context learning for inference and supervised fine-tuning for training. 
To demonstrate the \textit{effectiveness and efficiency} of our student models, we compare them against models from two categories: conventional bundle generation methods and LLM-based methods. For conventional methods, \textbf{Freq}~\cite{sun2022revisiting} employs the Apriori algorithm to identify frequent patterns at the item category level. \textbf{BBPR}~\cite{pathak2017generating} is a greedy algorithm that dynamically generates bundles by computing the similarity between users, items, and bundles representation learned by BPRMF~\cite{rendle2012bpr}. \textbf{POG}~\cite{chen2019pog} utilizes an encoder-decoder model based on Transformer architecture to generate personalized outfits by using multi-modal data. For LLM-based methods, \textbf{Zero-shot} method prompts GPT-3.5-turbo to generate bundles directly, which is our teacher model. \textbf{AICL}~\cite{sun2024adaptive} enhances Zero-shot by incorporating dynamic demonstration generation with retrieval augmentation to infer user intents and generate bundles simultaneously, establishing it as the state-of-the-art (SOTA) method. 
Specifically, comparisons with conventional methods aim to validate the superior effectiveness of our student models in generating high-quality bundles, while comparisons with LLM-based methods aim to assess whether they can deliver comparable performance with significantly enhanced efficiency.

\smallskip\noindent\textbf{Evaluation Metrics}. {We adopt the same evaluation metrics as in prior work~\cite{sun2022revisiting, sun2024revisiting} to evaluate the quality of generated bundles. Specifically, at the session level, we adopt \textit{Precision} and \textit{Recall} to measure the quantity of correctly predicted bundles within each session. \textit{Precision} indicates how many of the generated bundles are correct, while \textit{Recall} measures how many of the ground truth bundles are identified. Note that we treat a generated bundle as a hit bundle if it either completely matches or is a subset of a ground truth bundle. At the bundle level, we use \textit{Coverage} to evaluate the item-wise accuracy of hit bundles by calculating the average proportion of correctly predicted items within the ground truth bundles.}

\smallskip\noindent\textbf{Implement Details}.
For conventional bundle generation methods, we adopt the same parameter configurations suggested in~\cite{sun2024adaptive}. For LLM-based methods, we leverage the OpenAI API with GPT-3.5-turbo to implement various bundle generation strategies (Zero-shot and AICL) and to distill domain knowledge (rule and thought). Our student model Llama3.1-SFT is fine-tuned with QLoRA~\cite{dettmers2023qlora} based on the framework Unsloth\footnote{https://unsloth.ai/}. 
We conduct systematic hyperparameter tuning via grid search for learning rates in the range of \{2e-5, 8e-5, 2e-4\}, epochs in the range of \{3, 4, 5\}, QLoRA rank in the range of \{8, 16, 32\}, and QLoRA alpha values in the range of \{8, 16, 32\}. All experiments are conducted on 4× NVIDIA A40 GPUs with 48GB memory, with a batch size of 4 and gradient accumulation steps of 4.

\begin{table*}[t]
\centering
\small  
\caption{Performance comparison regarding different formats of distilled knowledge.}
\vspace{-0.1in}
\setlength{\tabcolsep}{4pt}  
\begin{tabular}{l|c|ccc|ccc|ccc}
\toprule
\multirow{2}{*}{Method} & \multirow{2}{*}{Knowledge} & \multicolumn{3}{c|}{Electronic} & \multicolumn{3}{c|}{Clothing} & \multicolumn{3}{c}{Food} \\
\cline{3-11}
& & Precision & Recall & Coverage & Precision & Recall & Coverage & Precision & Recall & Coverage \\
\midrule
\rowcolor{gray!30}
\multicolumn{2}{l|}{Zero-shot (Teacher)} & 0.580 & 0.820 & 0.720 & 0.603 & 0.752 & 0.788 & 0.604 & 0.815 & 0.748 \\\hline
\multirow{5}{*}{\begin{tabular}[c]{@{}l@{}}Llama3.1\\-ICL\end{tabular}} 
& Raw Data & 0.564 & 0.582 & 0.685 & 0.582 & 0.644 & 0.655 & 0.571 & 0.591 & 0.633 \\
& Pattern & 0.574 & 0.514 & 0.636 & 0.577 & 0.529 & 0.717 & 0.602 & 0.533 & 0.656 \\
& Rule & \textbf{0.611} & \textbf{0.615} & \textbf{0.693} & \textbf{0.621}* & \textbf{0.633}* & \textbf{0.768} & \textbf{0.621} & \textbf{0.601} & \textbf{0.773} \\
& Thought & 0.585 & 0.529 & 0.685 & 0.603 & 0.611 & 0.757 & 0.608 & 0.589 & 0.751 \\

\hline
\multirow{5}{*}{\begin{tabular}[c]{@{}l@{}}Llama3.1\\-SFT\end{tabular}}
& Raw Data & 0.618 & 0.621 & 0.857 & 0.608 & 0.607 & 0.901 & 0.615 & 0.597 & 0.841 \\
& Pattern & 0.626 & 0.615 & 0.818 & 0.611 & \textbf{0.623} & 0.860 & \textbf{0.668}* & \textbf{0.649}* & 0.842 \\
& Rule & 0.610 & 0.594 & \textbf{0.858}* & 0.609 & 0.598 & \textbf{0.915}* & 0.623 & 0.600 & \textbf{0.862}* \\
& Thought & \textbf{0.633}* & \textbf{0.644}* & 0.725 & \textbf{0.614} & 0.621 & 0.828 & 0.635 & 0.640 & 0.748 \\\hline
\multicolumn{2}{l|}{\textit{Improve}} &9.14\% &-21.5\% &19.17\% &2.99\% &-15.82\%&16.12\% &10.60\% &-20.37\% &15.24\% \\

\hline
\end{tabular}
\label{tab:rq1_res}
\vspace{-0.15in}
\end{table*}

\subsection{Investigation on Formats of Distilled Knowledge (RQ1)}\label{sec:results-rq1}

This section explores how different formats of distilled knowledge impact the bundle generation performance of student models, directly addressing RQ1. Table~\ref{tab:rq1_res} presents the experimental results across all metrics and domains, comparing the teacher model (first row) with two types of student models Llama3.1-ICL (rows 2-5) and Llama3.1-SFT (rows 6-9) using different types of knowledge, where the best performance achieved by both types of student models is in bold; and `Improve' (last row) indicates the relative improvements achieved by the best student model (marked by `*') compared with the teacher model\footnote{Unless otherwise specified, each type of knowledge (e.g., Pattern) used in the reported results refers to the accumulated knowledge from all three domains (e.g., Pattern mined from the three domains), ensuring a fair and consistent comparison.}. Several observations can be noted. 

First, \textit{for all student models, incorporating explicit knowledge — regardless of the type — leads to performance improvements to some extent. However, the most effective type of knowledge varies between student models with ICL and SFT}. 
Specifically, for Llama3.1-ICL, compared to using raw data (no explicit knowledge), Rule achieves the best performance across all domains and metrics, while Pattern and Thought yield better results in some cases. This suggests that rule knowledge is more acceptable for student LLM with ICL due to its ease of understanding and versatility. As for pattern knowledge, while it identifies frequent patterns in bundles, the category-level patterns tend to make the student LLM focus too much on relationships between categories while overlooking fine-grained item information. For example, if a session includes two cases for different tablet types and one screen protector, the LLM might incorrectly pair the protector with the wrong case, resulting in an inaccurate bundle. Although thought knowledge illustrates the reasoning process, in the ICL setting, relevant thoughts are retrieved from the training data based on semantic similarity to the target session, which may not perfectly align with the current session, thus leading to a performance drop. 
For Llama3.1-SFT, Thought achieves the highest Precision in the Electronic and Clothing domains, while Pattern yields the best Recall in both the Clothing and Food domains. In contrast, Rule demonstrates superior performance only in terms of Coverage. These results may be attributed to the following reasons: (1) injecting thought knowledge enhances the student LLM’s reasoning ability, thus identifying more precise bundles from the target session; and (2) injecting pattern knowledge helps the student LLM better identify and capture global trends, thereby improving Recall. 

Second, \textit{among the two types of student models, Llama3.1-SFT outperforms Llama3.1-ICL in most cases across all knowledge formats and metrics in the three domains.} 
This highlights the advantage of SFT, which explicitly trains the student model using distilled knowledge, enabling it to better internalize the task in the latent space and generate more accurate bundles. In contrast, ICL operates solely at inference time, leveraging distilled knowledge presented in the prompt without updating the model’s internal parameters. While ICL can dynamically retrieve relevant context, it lacks the capacity to fully internalize the task structure. Consequently, SFT's ability to align the model's internal representations with the bundle generation task results in more robust and reliable performance compared to the purely prompt-based approach of ICL. {Interestingly, in the Clothing domain, Llama3.1-ICL surpasses its Llama3.1-SFT on Precision and Recall, suggesting that with the right type of knowledge, ICL can still offer promising performance.}

Last, \textit{by integrating different formats of knowledge with various utilization methods, the student model can surpass the teacher model in terms of both Precision and Coverage; however, there remains a substantial gap in Recall (see row `Improve').}
This suggests that the student model fails to generate many of the relevant bundles that the teacher model identifies. The possible reasons are two-fold. 
(1) \textbf{Limited coverage of distilled knowledge}: The distilled knowledge often captures only the most common or easily formalizable aspects of bundle generation, potentially missing rare or complex cases that the teacher model is capable of recognizing. Consequently, guiding or training the student model with such generalized knowledge tends to improve the quality of typical bundles, resulting in higher Precision and Coverage, but fails to account for less frequent cases, thereby lowering Recall. (2) \textbf{Limited expressiveness of distilled knowledge}: When the teacher’s knowledge is distilled into explicit formats such as patterns, rules, or thoughts, the student model receives only high-level summaries or derivations rather than the full spectrum of the teacher’s latent understanding. The teacher model’s performance may hinge on complex and implicit reasoning paths that are difficult to capture through structured representations, which in turn limits the student model’s ability to recover all relevant bundles, further contributing to lower Recall.

Overall, the core answer to RQ1 is: 

\begin{answerbox}
\textit{Answer to RQ1:}
Different formats of distilled knowledge positively impact bundle generation performance by enhancing student models, and their effectiveness depends on the utilization methods (ICL vs. SFT). While Llama3.1-SFT benefits more consistently from different formats of distilled knowledge, even surpassing the teacher model in Precision and Coverage, challenges remain in achieving comparable Recall.
\end{answerbox}

\subsection{Investigation on Quantity of Distilled Knowledge (RQ2)}\label{sec:results-rq2} 

This section conducts extensive experiments, aiming to address RQ2: To what extent does the quantity of distilled knowledge influence bundle generation performance? 
To explore this, we employ the three approaches introduced in Section~\ref{sec:quantity-of-knowledge} to obtain varying amounts of distilled knowledge.
In particular, we first apply different data sampling strategies (i.e., random-based, length-based, diversity-based, and difficulty-based) with different sample ratios to capture varying distilled knowledge quantities.
Second, we aggregate the same type of knowledge from a single domain to multiple domains and evaluate the corresponding performance (e.g., Pattern from one domain vs. Pattern mined from three domains).
Third, we accumulate knowledge in different formats and compare the performance of using a single format versus combining multiple formats (e.g., Pattern vs. Pattern + Rule).

\input{chart/ratio_untuned_res_elec}
\input{chart/ratio_untuned_res_cloth}
\input{chart/ratio_untuned_res_food}

\subsubsection{Results on Sampling Strategies}\label{sec:results-sampling-strategies}
Figures~\ref{fig:ratio_untuned_res_elec}-\ref{fig:ratio_untuned_res_food} present the results on Llama3.1-ICL across the three domains on the three metrics. Some observations can be found. (1) \textit{More is better.} In most cases, the performance consistently scales with the sampling ratio across different sampling strategies. Higher ratio (e.g., 0.7 or 1.0) usually yield better results than lower ratios. This indicates that ICL's effectiveness relies heavily on access to a large volume of knowledge. A richer knowledge pool provides more relevant examples for the retrieval mechanism to construct effective prompts. (2) \textit{The impact of different sampling strategies is limited.} The performance gap between different sampling strategies at the same ratio is relatively minor, suggesting that the quantity of retrievable data is more critical than the specific method used to select it.
However, a counter-intuitive observation arises when examining Recall with Rule and Thought: increasing the sample ratio does not necessarily lead to improved Recall in the Electronic and Food domains. A possible explanation is the high bundle diversity in these domains compared to Clothing. For example, Electronics bundles can range from computer-related items to camera gear, while Food bundles vary from snacks to full meals. In contrast, the Clothing domain exhibits more homogeneity, making it easier for fine-grained knowledge (i.e., Rule and Thought) to help the student model group similar items effectively. In the case of Electronics and Food, this detailed knowledge may cause the model to overfit to specific bundle types, narrowing its focus and reducing its ability to generate diverse valid combinations. As a result, the total number of generated bundles decreases, leading to lower overall Recall.
Besides, for Pattern knowledge, its trend in Coverage differs from other types of knowledge. This is mainly because frequent patterns typically involve only 2 to 4 categories, which encourages the student model to generate shorter bundles, ultimately lowering Coverage.

\input{chart/ratio_tuned_elec}
\input{chart/ratio_tuned_cloth}

Figures~\ref{fig:ratio_tuned_elec}-\ref{fig:ratio_tuned_food} illustrate the performance of Llama3.1-SFT across the three domains, revealing different observations compared to Llama3.1-ICL. (1) \textit{The performance improves as the sampling ratio (i.e., knowledge) increases, reaching its peak with a certain ratio (e.g., 0.7) and then shows a slight drop}. This suggests that for Llama3.1-SFT, sampling data at an appropriate quantity range can achieve optimal performance, while adding more data yields limited benefits even degradation. 
(2) \textit{Sampling strategy has a greater impact on performance}. Different sampling strategies lead to relatively noticeable performance difference at the same sampling ratio. Specifically, for Llama3.1-ICL, the average performance standard deviation of the four sampling strategies across different sample ratios, knowledge formats, and domains for Precision, Recall and Coverage are {0.013, 0.013, 0.01}, respectively. In contrast, the corresponding average performance standard deviation for Llama3.1-SFT are {0.025, 0.029, 0.022}. The difference is statistically significant, as confirmed by a paired t-test with a $p$-value < 0.01. These results suggest that Llama3.1-SFT is more sensitive to the choice of sampling strategy compared to Llama3.1-ICL.

\input{chart/ratio_tuned_food}
\input{chart/res_domain_scale}

\subsubsection{Results on Knowledge Accumulation from Different Domains}
First, we find that \textit{accumulation from different domains does not necessarily impact the performance of Llama3.1-ICL.} This is because ICL typically retrieves the session most similar to the target session and utilizes its associated knowledge to enhance bundle generation performance. However, sessions from other domains are often dissimilar to the target session, making their accumulated knowledge less likely to be leveraged.
Second, we observe that \textit{aggregating knowledge from multiple domains consistently outperforms using knowledge from a single domain, regardless of the knowledge type across the three domains when using Llama3.1-SFT}. In particular, Figure~\ref{fig:domain_scale} illustrates the performance comparison between accumulating different knowledge types from single (blue line) to multiple domains (red line) using Llama3.1-SFT, where the x-axis represents different types of knowledge, and "Raw" indicates that no explicit knowledge is considered. This improvement can be attributed to the nature of SFT: unlike ICL, SFT leverages the accumulated knowledge to fine-tune the student model. Aggregated knowledge from multiple domains (1) brings diversity, preventing the model from over-fitting; and (2) provides complementary insights, allowing the model to learn more generalizable patterns for bundle generation. For instance, understanding the concept of `complementary accessories' from Electronic might help generalize to `coordinating items' in Clothing, even if the specific items differ.

\subsubsection{Results on Knowledge Accumulation from Different Formats}\label{sec:result-formats}
Figures~\ref{fig:heatmaps_ICL} and ~\ref{fig:heatmaps_SFT} present the results of accumulating different types of knowledge in the three domains, where y-axis represents different combinations of knowledge (from single type to two and three types). Some observations can be obtained. 

(1) \textit{For Llama3.1-ICL, combining two types of knowledge (e.g., Pattern + Rule or Rule + Thought) generally improves Precision and Coverage across all three domains compared to using a single knowledge type, but does not improve Recall}. The gain in Precision and Coverage may stem from the model receiving more comprehensive guidance through multiple knowledge types, thus generating more precise bundles. 
However, using Rule alone yields the highest Recall in the Electronic and Clothing domains, and a relatively high Recall in the Food domain. This suggests that, in the ICL setting, adding more knowledge types may increase the model’s confidence in identifying the most relevant bundle within a session. As a result, the model becomes more conservative, focusing on a small number of high-confidence predictions while overlooking other potential candidates. Such behavior increases Precision at the expense of Recall.
Moreover, we notice that integrating all types of knowledge does not lead to further improvements. This is likely due to two main challenges: (i) the longer context increases the model’s processing burden, and (ii) potential inconsistencies among different knowledge formats may introduce confusion. Compared to combining two knowledge types, merging all three increases the risk of conflicting signals, making it harder for the model to determine which guidance to prioritize, ultimately leading to diminished performance.

(2) \textit{For Llama3.1-SFT, combining all three types of knowledge yields the highest Precision and Recall in the Electronic and Clothing domains. Unlike Llama3.1-ICL, the performance trends for Precision and Recall with Llama3.1-SFT are more consistent}. This can be attributed to the nature of SFT, which updates the model’s internal parameters through paired input-output examples. This fine-tuning process enables the model to effectively integrate diverse signals and understand how different types of knowledge interact to support bundle generation. As a result, the model can leverage the complementary strengths of various knowledge formats, leading to improvements in both Precision and Recall, as well as greater stability in performance. However, the model's ability to balance diverse knowledge sources can still be improved. Notably, the best Coverage is achieved by combining only two knowledge types — Pattern and Rule — suggesting that the integration of all three knowledge types may still introduce redundant or conflicting signals, thereby limiting the model’s capacity to generate bundles of greater size.

\pgfplotsset{colormap={blues}{
    rgb255(0.4)=(222,235,247);
    rgb255(0.6)=(158,202,225);
    rgb255(0.8)=(49,130,189);
    rgb255(1.0)=(8,48,107)
}}

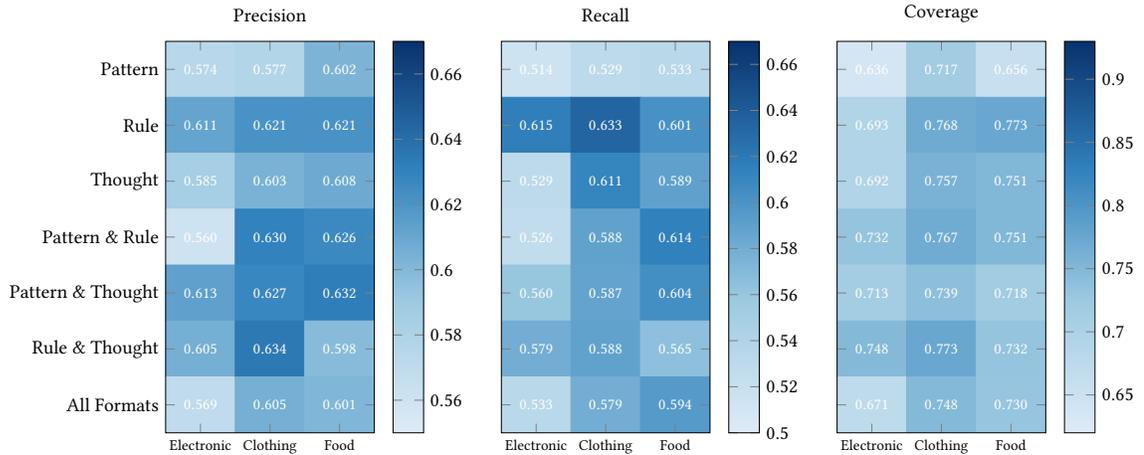
\begin{figure}[tbp]
\centering
\begin{adjustbox}{max width=\textwidth}
\begin{tikzpicture}
  \begin{axis}[title={Precision},
    ylabel={},
    xtick={1,2,3},
    point meta min=0.55,
    point meta max=0.67,
    xticklabels={Electronic,Clothing,Food},
    xticklabel style={font=\footnotesize},
    ytick={1,2,3,4,5,6,7},
    yticklabels={Pattern,Rule,Thought,Pattern \& Rule,Pattern \& Thought,Rule \& Thought,All Formats},
    enlargelimits=false,
    colorbar,
    colormap name=blues,
    width=5cm,height=8cm
  ]
    \addplot[
      matrix plot,
      point meta=explicit,
      mesh/cols=3,
      mesh/rows=7,
      nodes near coords,
      nodes near coords align={center},
      nodes near coords style={
        font=\footnotesize,
        text=white
      },
      every node near coord/.append style={
        /pgf/number format/.cd,
        fixed,
        precision=3,
        zerofill
      }
    ] coordinates {
      (1,1) [0.574] (2,1) [0.577] (3,1) [0.602]
      (1,2) [0.611] (2,2) [0.621] (3,2) [0.621]
      (1,3) [0.585] (2,3) [0.603] (3,3) [0.608]
      (1,4) [0.560] (2,4) [0.630] (3,4) [0.626]
      (1,5) [0.613] (2,5) [0.627] (3,5) [0.632]
      (1,6) [0.605] (2,6) [0.634] (3,6) [0.598]
      (1,7) [0.569] (2,7) [0.605] (3,7) [0.601]
    };
  \end{axis}
  \begin{axis}[title={Recall}, 
    yticklabels={},
    xtick={1,2,3},
    xticklabels={Electronic,Clothing,Food},
    xticklabel style={font=\footnotesize},
    ytick={},
    point meta min=0.5,
    point meta max=0.67,
    enlargelimits=false,
    at={(5.5cm,0cm)},
    colorbar,
    colormap name=blues,
    width=5cm,height=8cm
  ]
    \addplot[
      matrix plot,
      point meta=explicit,
      mesh/cols=3,
      mesh/rows=7,
      nodes near coords,
      nodes near coords align={center},
      nodes near coords style={
        font=\footnotesize,
        text=white
      },
      every node near coord/.append style={
        /pgf/number format/.cd,
        fixed,
        precision=3,
        zerofill
      }
    ] coordinates {
      (1,1) [0.514] (2,1) [0.529] (3,1) [0.533]
      (1,2) [0.615] (2,2) [0.633] (3,2) [0.601]
      (1,3) [0.529] (2,3) [0.611] (3,3) [0.589]
      (1,4) [0.526] (2,4) [0.588] (3,4) [0.614]
      (1,5) [0.560] (2,5) [0.587] (3,5) [0.604]
      (1,6) [0.579] (2,6) [0.588] (3,6) [0.565]
      (1,7) [0.533] (2,7) [0.579] (3,7) [0.594]
    };
  \end{axis}
  \begin{axis}[title={Coverage},
    yticklabels={},
    xtick={1,2,3},
    xticklabels={Electronic,Clothing,Food},
    xticklabel style={font=\footnotesize},
    ytick={},
    point meta min=0.62,
    point meta max=0.93,
    enlargelimits=false,
    at={(11cm,0cm)},
    colorbar,
    colormap name=blues,
    width=5cm,height=8cm,
  ]
    \addplot[
      matrix plot,
      point meta=explicit,
      mesh/cols=3,
      mesh/rows=7,
      nodes near coords,
      nodes near coords align={center},
      nodes near coords style={
        font=\footnotesize,
        text=white
      },
      every node near coord/.append style={
        /pgf/number format/.cd,
        fixed,
        precision=3,
        zerofill
      }
    ] coordinates {
      (1,1) [0.636] (2,1) [0.717] (3,1) [0.656]
      (1,2) [0.693] (2,2) [0.768] (3,2) [0.773]
      (1,3) [0.692] (2,3) [0.757] (3,3) [0.751]
      (1,4) [0.732] (2,4) [0.767] (3,4) [0.751]
      (1,5) [0.713] (2,5) [0.739] (3,5) [0.718]
      (1,6) [0.748] (2,6) [0.773] (3,6) [0.732]
      (1,7) [0.671] (2,7) [0.748] (3,7) [0.730]
    };
  \end{axis}
\end{tikzpicture}
\end{adjustbox}
\caption{Results of knowledge accumulation from different formats in the Electronic, Clothing, and Food domains using Llama3.1-ICL.}
\label{fig:heatmaps_ICL}
\vspace{-0.15in}
\end{figure}
\pgfplotsset{colormap={blues}{
    rgb255(0.4)=(222,235,247);
    rgb255(0.6)=(158,202,225);
    rgb255(0.8)=(49,130,189);
    rgb255(1.0)=(8,48,107)
}}

\begin{figure}[tbp]
\centering
\begin{adjustbox}{max width=\textwidth}
\begin{tikzpicture}
  \begin{axis}[title={Precision},
    ylabel={},
    xtick={1,2,3},
    xticklabels={Electronic,Clothing,Food},
    xticklabel style={font=\footnotesize},
    ytick={1,2,3,4,5,6,7},
    point meta min=0.55,
    point meta max=0.67,
    yticklabels={Pattern,Rule,Thought,Pattern \& Rule,Pattern \& Thought,Rule \& Thought,All Formats},
    enlargelimits=false,
    colorbar,
    colormap name=blues,
    width=5cm,height=8cm
  ]
    \addplot[
      matrix plot,
      point meta=explicit,
      mesh/cols=3,
      mesh/rows=7,
      nodes near coords,
      nodes near coords align={center},
      nodes near coords style={
        font=\footnotesize,
        text=white
      },
      every node near coord/.append style={
        /pgf/number format/.cd,
        fixed,
        precision=3,
        zerofill
      }
    ] coordinates {
      (1,1) [0.626] (2,1) [0.611] (3,1) [0.668]
      (1,2) [0.610] (2,2) [0.609] (3,2) [0.623]
      (1,3) [0.633] (2,3) [0.614] (3,3) [0.635]
      (1,4) [0.610] (2,4) [0.638] (3,4) [0.647]
      (1,5) [0.582] (2,5) [0.646] (3,5) [0.627]
      (1,6) [0.619] (2,6) [0.619] (3,6) [0.648]
      (1,7) [0.642] (2,7) [0.665] (3,7) [0.629]
    };
  \end{axis}
  \begin{axis}[title={Recall}, 
    yticklabels={},
    xtick={1,2,3},
    xticklabels={Electronic,Clothing,Food},
    xticklabel style={font=\footnotesize},
    ytick={},
    point meta min=0.5,
    point meta max=0.67,
    enlargelimits=false,
    at={(5.5cm,0cm)},
    colorbar,
    colormap name=blues,
    width=5cm,height=8cm
  ]
    \addplot[
      matrix plot,
      point meta=explicit,
      mesh/cols=3,
      mesh/rows=7,
      nodes near coords,
      nodes near coords align={center},
      nodes near coords style={
        font=\footnotesize,
        text=white
      },
      every node near coord/.append style={
        /pgf/number format/.cd,
        fixed,
        precision=3,
        zerofill
      }
    ] coordinates {
      (1,1) [0.615] (2,1) [0.623] (3,1) [0.659]
      (1,2) [0.594] (2,2) [0.598] (3,2) [0.600]
      (1,3) [0.644] (2,3) [0.625] (3,3) [0.640]
      (1,4) [0.598] (2,4) [0.587] (3,4) [0.630]
      (1,5) [0.634] (2,5) [0.618] (3,5) [0.651]
      (1,6) [0.640] (2,6) [0.620] (3,6) [0.639]
      (1,7) [0.656] (2,7) [0.625] (3,7) [0.595]
    };
  \end{axis}
  \begin{axis}[title={Coverage},
    yticklabels={},
    xtick={1,2,3},
    point meta min=0.62,
    point meta max=0.93,
    xticklabels={Electronic,Clothing,Food},
    xticklabel style={font=\footnotesize},
    ytick={},
    enlargelimits=false,
    at={(11cm,0cm)},
    colorbar,
    colormap name=blues,
    width=5cm,height=8cm,
  ]
    \addplot[
      matrix plot,
      point meta=explicit,
      mesh/cols=3,
      mesh/rows=7,
      nodes near coords,
      nodes near coords align={center},
      nodes near coords style={
        font=\footnotesize,
        text=white
      },
      every node near coord/.append style={
        /pgf/number format/.cd,
        fixed,
        precision=3,
        zerofill
      }
    ] coordinates {
      (1,1) [0.818] (2,1) [0.860] (3,1) [0.842]
      (1,2) [0.858] (2,2) [0.915] (3,2) [0.862]
      (1,3) [0.725] (2,3) [0.828] (3,3) [0.748]
      (1,4) [0.871] (2,4) [0.920] (3,4) [0.826]
      (1,5) [0.726] (2,5) [0.855] (3,5) [0.762]
      (1,6) [0.715] (2,6) [0.804] (3,6) [0.766]
      (1,7) [0.770] (2,7) [0.832] (3,7) [0.764]
    };
  \end{axis}
\end{tikzpicture}
\end{adjustbox}
\caption{Results of knowledge accumulation from different formats in the Electronic, Clothing, and Food domains using Llama3.1-SFT.}
\label{fig:heatmaps_SFT}
\vspace{-0.15in}
\end{figure}
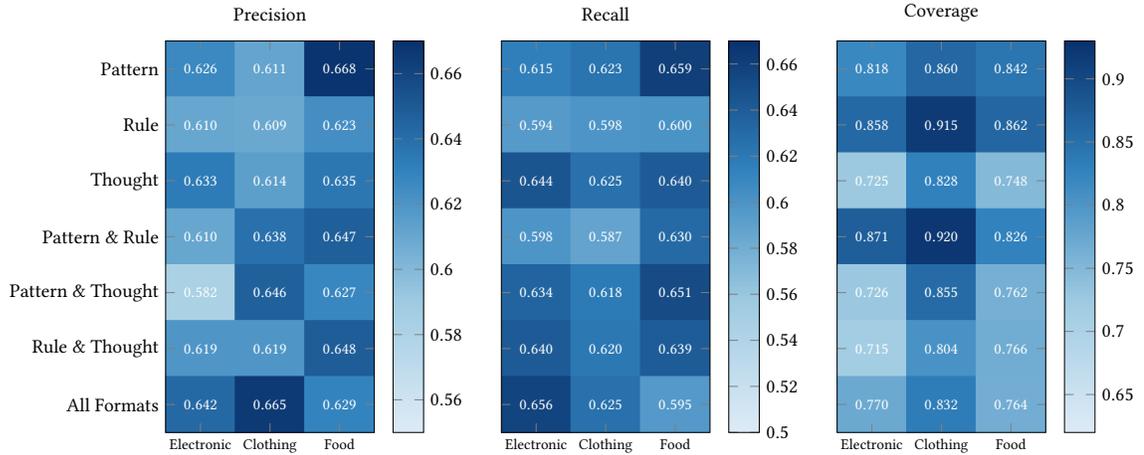

According to the analysis in Sections~\ref{sec:results-sampling-strategies}-\ref{sec:result-formats}, the answer to RQ2 is: 
\begin{answerbox}
\textit{Answer to RQ2:} Overall, increasing the quantity of distilled knowledge positively impacts bundle generation performance. Specifically, ICL benefits more from higher sampling ratios, as it relies on a larger knowledge pool for effective retrieval. In contrast, SFT tends to peak at a certain sampling ratio and gains substantial improvements from incorporating knowledge across different formats and domains.
\end{answerbox}

\input{chart/stage_res_diff_multi_domain}

\subsection{Investigation on Utilization Method of Distilled Knowledge (RQ3)}\label{sec:results-rq3}

In this section, we design experiments to investigate the impact of different ways of utilizing the distilled knowledge on bundle generation to answer RQ3. Specifically, three methods are considered, including ICL, SFT and ICL+SFT, as introduced in Section~\ref{sec:utilizing-methods}. 
Figure~\ref{fig:merged_stage_res_diff_multi} shows the results of applying the three methods to the student LLM (Llama3.1-8B) with different knowledge.  
{In these figures, the colored bars (SFT+K) represent SFT with different types of distilled knowledge K for fine-tuning (e.g., the blue bar represents using Pattern knowledge in the SFT); the x-axis (ICL+K) represents ICL with different types of distilled knowledge K for inference (e.g., ICL+Pattern means Pattern is used in ICL); and the y-axis represents the performance differences (gaps) between (ICL+K) and different combinations of (ICL+K) and (SFT+K)\footnote{We compute the y-value by subtracting the performance of (ICL+K) from that of each combination (ICL+K, SFT+K).}. For instance, the blue bar under ICL+Pattern represents the performance difference between ICL+Pattern and (ICL+Pattern, SFT+Pattern). Particularly, the purple bars show the performance of integrating knowledge only in the SFT, so the corresponding y-value indicates the performance difference of using the same type of distilled knowledge in ICL for inference and in SFT for fine-tuning (i.e., ICL+K vs. SFT+K).



From the results, four major observations can be noted. 
(1) \textit{The combination (ICL+K, SFT+K) generally yields better performance than either ICL+K or SFT+K alone, indicating that leveraging knowledge in both ICL and SFT is more effective than using it in only one of them}. As observed, in the majority of cases, the colored bars correspond to positive y-values, and bars in blue, red, green, and yellow exhibit higher y-values compared to those in purple. This suggests that the student model benefits from using the distilled knowledge both for fine-tuning its parameters and as guidance during inference. Alternatively stated, the student model fine-tuned with distilled knowledge can continue to benefit from domain-specific guidance during inference, further boosting task performance. 
(2) \textit{The combination (ICL+K, SFT+K) does not always achieve positive gains, which is highly dependent on the specific knowledge K used}. For instance, SFT+Pattern (blue bar) usually shows positive y-values, regardless of the type of knowledge used in ICL. Similarly, ICL+Pattern generally gains positive y-values regardless of the knowledge utilized in SFT. In contrast, the combination (ICL+Thought/All, SFT+Thought/All) tends to result in negative y-values in most cases. {This indicates a potential risk of combining divergent knowledge types across SFT and ICL: when the information introduced at each stage is not well-aligned, it can impair the model’s ability to focus and generalize effectively.}
(3) \textit{SFT+K alone yields positive y-values in most cases, suggesting that incorporating distilled knowledge into SFT is more effective than doing so in ICL (can also refer to Table~\ref{tab:rq1_res})}. Moreover, SFT can even outperform the combination (ICL+K, SFT+K) in certain cases — for example, in terms of Precision and Recall in the Electronic domain. This suggests a stronger ability of SFT in injecting knowledge into the student model.

Overall, the main answer to RQ3 is:

\begin{answerbox}
\textit{Answer to RQ3:} Bundle generation performance significantly depends on the knowledge utilization method. Using distilled knowledge in both SFT and ICL generally yields the best performance, but its effectiveness depends on the specific knowledge used. Notably, SFT with distilled knowledge alone is more consistently effective than ICL alone.
\end{answerbox}

\subsection{Comparison on Effectiveness and Efficiency}

The previous analysis in Sections~\ref{sec:results-rq1}-\ref{sec:results-rq3} demonstrates how the format, quantity, and utilization method of distilled knowledge affect the performance of the student model on the bundle generation task. Now, we further analyze the effectiveness and efficiency of the student model in comparison with both conventional models and its teacher models.

\subsubsection{{Analysis on Effectiveness}}
Table~\ref{tab:baselines} presents the performance of conventional models, teacher models, and student models across all metrics and domains. For student models, we report the best-performing result under each research question (denoted as Student\_RQX). Specifically, for RQ1, we select Llama3.1-SFT with Thought for the Electronic and Clothing domains, and Llama3.1-SFT with Pattern for the Food domain. For RQ2, the best results come from Llama3.1-SFT with Thought using diversity-based sampling (sampling ratio=0.7) for Electronic, Llama3.1-SFT with All formats of knowledge using full raw data for Clothing, and Llama3.1-SFT with Pattern using difficulty-based sampling (sampling ratio=0.7) for Food. For RQ3, we choose Llama3.1-SFT with All formats of knowledge in fine-tuning and ICL with Rule in inference for Electronic; Llama3.1-SFT with All formats of knowledge and ICL with Pattern for Clothing; and Llama3.1-SFT with Pattern and ICL with Thought for Food. The performance of the teacher model is highlighted in gray; the best performance achieved by the student models is in bold; and `Improve' indicates the relative improvements achieved by the best-performing student model against the teacher model.

\begin{table*}[t]
\centering
\small  
\caption{Effectiveness comparison across different models in the three domains.}
\vspace{-0.1in}
\setlength{\tabcolsep}{4pt}  
\begin{tabular}{l|c|ccc|ccc|ccc}
\toprule
\multirow{2}{*}{Type} & \multirow{2}{*}{Method} & \multicolumn{3}{c|}{Electronic} & \multicolumn{3}{c|}{Clothing} & \multicolumn{3}{c}{Food} \\
\cline{3-11}
& & Precision & Recall & Coverage & Precision & Recall & Coverage & Precision & Recall & Coverage \\
\midrule
\multirow{4}{*}{\begin{tabular}[c]{@{}l@{}}Conventional\\ Models\end{tabular}} & Freq & 0.423 & 0.597 & 0.701 & 0.532 & 0.566 & 0.698 & 0.491 & 0.525 & 0.684 \\
 & BBPR & 0.260 & 0.122 & 0.433 & 0.239 & 0.211 & 0.449 & 0.210 & 0.183 & 0.416 \\
 & POG & 0.339 & 0.250 & 0.412 & 0.312 & 0.221 & 0.399 & 0.365 & 0.266 & 0.393 \\
 & BYOB & 0.340 & 0.294 & 0.361 & 0.311 & 0.273 & 0.457 & 0.304 & 0.253 & 0.427 \\
\hline
\multirow{5}{*}{\begin{tabular}[c]{@{}l@{}}LLM-based\\Models\end{tabular}}& AICL & 0.769 &0.859 & 0.741 & 0.677 & 0.788 & 0.839 & 0.698 & 0.851 & 0.755 \\
&\cellcolor{gray!30}Teacher Model & \cellcolor{gray!30}0.580 & \cellcolor{gray!30}0.820 & \cellcolor{gray!30}0.720 & \cellcolor{gray!30}0.603 & \cellcolor{gray!30}0.752 & \cellcolor{gray!30}0.788 & \cellcolor{gray!30}0.604 & \cellcolor{gray!30}0.815 & \cellcolor{gray!30}0.748 \\
& Student\_RQ1 & 0.633 & 0.644 & 0.725 & 0.614 & 0.621 & 0.828 & 0.668 & 0.649 & \textbf{0.862} \\
& Student\_RQ2 & 0.650 & \textbf{0.682} & \textbf{0.783} & 0.665 & \textbf{0.625} & 0.832 & \textbf{0.707} & 0.681 & 0.834 \\
& Student\_RQ3 & \textbf{0.669} & 0.626 & 0.761 & \textbf{0.677} & 0.606 & \textbf{0.840} & 0.704 & \textbf{0.685} & 0.850 \\
\hline
\multicolumn{2}{l|}{\textit{Improve}} &15.35\% &-16.83\% &8.75\% &12.27\% &-16.89\%&6.60\% &17.05\% &-15.95\% &15.24\% \\
\bottomrule
\end{tabular}\label{tab:baselines}
\vspace{-0.15in}
\end{table*}

We make the following observations. 
(1) \textit{LLMs-based models outperform the conventional models across all metrics.} Both the teacher and student models based on LLMs, consistently achieve better performance across all metrics and domains compared to conventional models. This advantage stems from their extensive pre-training, inherent reasoning capabilities, and effective adaptation even with limited task-specific data—a scenario where conventional models may be underfitting.
(2) \textit{Knowledge distillation enables student models to outperform the teacher model on certain metrics.} While the teacher model achieves the best Recall, knowledge distillation successfully empowers the smaller student model to achieve highly competitive, and in some aspects superior, results. In particular, the best-performing student models consistently outperform the teacher model in Precision and Coverage across all domains. Additionally, when compared to AICL, the advanced LLM-based method for bundle generation, the student models achieve higher Coverage across the three domains, and notably higher Precision in the Food domain. These results strongly confirm the effectiveness of knowledge distillation for more effective bundle generation. 
(3) \textit{Optimizing knowledge utilization method (RQ3) is crucial for maximizing the performance of student models.} As shown in Table~\ref{tab:impact-of-factor}, we calculate the relative performance gap under each research question across the three domains, defined as $(best\_result-worst\_result)/worst\_result$\footnote{We report the result for each domain by averaging the relative performance gaps across the three metrics (Precision, Recall, and Coverage). For RQ1 and RQ2, we also compute the gaps separately for each utilization method (i.e., ICL and SFT), and then average the gaps of the two methods to obtain an overall result for each RQ.}. 
Intuitively, larger performance gaps reflect stronger impacts. Based on the results, we find the following observations. (i) Among the three factors (format, quantity, and utilization method), the utilization method generally has the greatest overall impact on model performance, while the knowledge format contributes the least. This trend is evident in the results with blue background - RQ1: 11.33\%, RQ2: 23.21\%, RQ3: 39.62\%. (ii) This pattern is consistent across individual domains (highlighted with red backgrounds), with the exception of the Food domain. In that case, knowledge quantity shows a slightly higher impact than the utilization method (RQ2: 24.84\% vs. RQ3: 24.58\%), though the difference is marginal.
(iii) For both ICL and SFT, knowledge quantity consistently has a stronger influence than knowledge format. This effect is more pronounced in SFT. Specifically, as shown in the orange background for ICL: RQ1: 13.90\% vs. RQ2: 18.93\%; and in the green background for SFT: RQ1: 8.75\% vs. RQ2: 27.48\%.

\subsubsection{Analysis on Efficiency} Due to limited resources, we were unable to deploy the teacher and student models in identical runtime environments, which makes it difficult to fairly quantify the inference latency between the teacher and student models. Therefore, as an alternative measure of efficiency, we report the computational resources consumed by the teacher and student models during fine-tuning and inference, respectively. Due to the scale difference between the teacher and student models, there is a significant gap in resource consumption. The teacher model (e.g., GPT-3.5-turbo)
requires around 1450GB of memory for full fine-tuning and 250GB for inference, often relying on multi-GPU setups and advanced optimization strategies. In contrast, the student model (e.g., Llama3.1-8B) only requires about 122GB for full fine-tuning and 20GB for inference, which can be comfortably handled on a single A100 or even a consumer-grade RTX 3090. With parameter-efficient tuning methods like LoRA~\cite{dettmers2023qlora} or quantization techniques~\cite{li2024quantized}, the student model’s resource footprint can be further reduced to as low as 20–30GB during training and 4–8GB during inference, making it highly suitable for more accessible and cost-effective deployment.

\begin{table}[t]
    \centering
    \caption{The relative performance gap between the best and worst result under each research question across the three domains. The different background colors represent the influence of three factors—knowledge format, quantity, and utilization method—from various perspectives. Specifically, red, orange, and green indicate their impacts on different domains, ICL, and SFT, respectively, while blue reflects their overall impact. A darker shade corresponds to a larger performance gap, indicating a stronger influence.  
    }
    \label{tab:impact-of-factor}
    \vspace{-0.1in}
    \begin{tabular}{l|ccc|ccc|c}
    \toprule
    &\multicolumn{3}{c|}{RQ1: Format of Knowledge}   &\multicolumn{3}{c|}{RQ2: Quantity of Knowledge}  &\multicolumn{1}{c}{RQ3: Utilization Methods of Knowledge}  \\\cline{2-8}
    &ICL &SFT &Avg. &ICL &SFT &Avg. &Avg.\\\midrule
    Electronic &12.31\% &10.18\% &\cellcolor{red!13!white}11.25\% &22.54\% &28.11\% &\cellcolor{red!43!white}25.33\% &\cellcolor{red!73!white}44.56\%\\
    Clothing &14.85\% &5.22\% &\cellcolor{red!13!white}10.04\% &15.00\% &23.91\% &\cellcolor{red!43!white}19.46\% &\cellcolor{red!73!white}49.72\%\\
    Food &14.54\% &10.86\% &\cellcolor{red!13!white}12.70\% &19.25\% &30.42\% &\cellcolor{red!73!white}24.84\% &\cellcolor{red!43!white}24.58\%\\\hline
    Avg. &\cellcolor{orange!15!white}13.90\% &\cellcolor{green!13!white}8.75\% &\cellcolor{blue!13!white}11.33\% &\cellcolor{orange!45!white}18.93\% &\cellcolor{green!43!white}27.48\% &\cellcolor{blue!43!white}23.21\% &\cellcolor{blue!73!white}39.62\%\\\bottomrule
    \end{tabular}
    \vspace{-0.1in}
\end{table}

\section{Conclusion and Future Work}
In this work, we systematically explore knowledge distillation techniques for LLM-based bundle generation task, aiming to reduce the significant computational costs associated with large models while preserving their effectiveness. We propose a comprehensive KD framework featuring progressive knowledge extraction (frequent patterns, formalized rules, deep thoughts), diverse strategies to vary knowledge quantity (sampling, domain/format accumulation), and complementary LLM adaptation techniques (ICL, SFT and SFT+ICL) along with their combination for knowledge utilization.
Our extensive experiments across the three real-world bundle datasets demonstrate that knowledge distillation is indeed a viable and potent strategy. The effectiveness is nuanced, depending significantly on the interplay between the format of distilled knowledge, its quantity, and the method used to utilize it, as well as the specific characteristics of the dataset domain. To be specific, SFT with distilled knowledge consistently emerged as a strong approach, enabling smaller student models to achieve performance comparable, and on certain metrics (Precision, Coverage) even superior, to the large teacher model. The combined SFT+ICL approach often yields the best results, depending on careful selection of knowledge for each stage. Importantly, these results are obtained with significantly lower computational cost compared to the teacher LLM. Overall, our findings highlight the potential of efficient LLM-based solutions for complex tasks like bundle generation and offer practical insights for optimizing knowledge distillation strategies in future work.

\textbf{Limitations and Future Work.} While this work demonstrates the effectiveness of explicit knowledge distillation, it primarily focuses on transferring knowledge expressible in textual formats (patterns, rules, thoughts). This explicit approach may not fully capture the rich, implicit knowledge residing within the internal states (e.g., hidden representations, attention weights) of the teacher LLM. Therefore, a key direction for future work is to explore implicit knowledge distillation techniques, aiming to align the student model's internal processing more closely with the teacher's, potentially capturing more subtle reasoning patterns. Furthermore, our experiments revealed that the selection and combination of different knowledge formats and quantities significantly impact performance, especially for the combined SFT+ICL utilization method. Simply using all available knowledge is not always optimal. This highlights the need for more sophisticated knowledge selection and fusion mechanisms. Future research could focus on developing adaptive or model-driven methods to automatically determine the most beneficial knowledge subset or combination for a given task, domain, or utilization strategy, thereby improving the robustness and maximizing the potential of the student model. Finally, extending these knowledge distillation techniques to incorporate multi-modal data (e.g., item images) common in bundle generation scenarios remains an important avenue for enhancing effectiveness.

\begin{acks}
This paper is supported by Open Foundation of Key Laboratory of Interdisciplinary Research of Computation and Economics (Shanghai University of Finance and Economics), Ministry of Education, China. It is partially supported by the Ministry of Education, Singapore, under its MOE AcRF Tier 1, SUTD Kickstarter Initiative (SKI 2021\_06\_12). 
We also greatly acknowledge the support of the National Natural Science Foundation of China (Grant No. 72371148 and 72192832), the Shanghai Rising-Star Program (Grant No. 23QA1403100), and the Program for Innovative Research Team of Shanghai University of Finance and Economics.
Additionally, this paper has been refined by Claude and ChatGPT to enhance readability. 
\end{acks}

\bibliographystyle{ACM-Reference-Format}


\begin{thebibliography}{71}


\ifx \showCODEN    \undefined \def \showCODEN     #1{\unskip}     \fi
\ifx \showDOI      \undefined \def \showDOI       #1{#1}\fi
\ifx \showISBNx    \undefined \def \showISBNx     #1{\unskip}     \fi
\ifx \showISBNxiii \undefined \def \showISBNxiii  #1{\unskip}     \fi
\ifx \showISSN     \undefined \def \showISSN      #1{\unskip}     \fi
\ifx \showLCCN     \undefined \def \showLCCN      #1{\unskip}     \fi
\ifx \shownote     \undefined \def \shownote      #1{#1}          \fi
\ifx \showarticletitle \undefined \def \showarticletitle #1{#1}   \fi
\ifx \showURL      \undefined \def \showURL       {\relax}        \fi
\providecommand\bibfield[2]{#2}
\providecommand\bibinfo[2]{#2}
\providecommand\natexlab[1]{#1}
\providecommand\showeprint[2][]{arXiv:#2}

\bibitem[Agrawal et~al\mbox{.}(1994)]%
        {agrawal1994fast}
\bibfield{author}{\bibinfo{person}{Rakesh Agrawal}, \bibinfo{person}{Ramakrishnan Srikant}, {et~al\mbox{.}}} \bibinfo{year}{1994}\natexlab{}.
\newblock \showarticletitle{Fast algorithms for mining association rules}. In \bibinfo{booktitle}{\emph{Proceedings of 20th International Conference on Very Large Scale Data Bases (VLDB)}}, Vol.~\bibinfo{volume}{1215}. \bibinfo{pages}{487--499}.
\newblock


\bibitem[Avny~Brosh et~al\mbox{.}(2022)]%
        {avny2022bruce}
\bibfield{author}{\bibinfo{person}{Tzoof Avny~Brosh}, \bibinfo{person}{Amit Livne}, \bibinfo{person}{Oren Sar~Shalom}, \bibinfo{person}{Bracha Shapira}, {and} \bibinfo{person}{Mark Last}.} \bibinfo{year}{2022}\natexlab{}.
\newblock \showarticletitle{BRUCE: Bundle recommendation using contextualized item embeddings}. In \bibinfo{booktitle}{\emph{Proceedings of the 16th ACM Conference on Recommender Systems (RecSys)}}. \bibinfo{pages}{237--245}.
\newblock


\bibitem[Bai et~al\mbox{.}(2019)]%
        {bai2019personalized}
\bibfield{author}{\bibinfo{person}{Jinze Bai}, \bibinfo{person}{Chang Zhou}, \bibinfo{person}{Junshuai Song}, \bibinfo{person}{Xiaoru Qu}, \bibinfo{person}{Weiting An}, \bibinfo{person}{Zhao Li}, {and} \bibinfo{person}{Jun Gao}.} \bibinfo{year}{2019}\natexlab{}.
\newblock \showarticletitle{Personalized bundle list recommendation}. In \bibinfo{booktitle}{\emph{The Web Conference (TheWebConf)}}. \bibinfo{pages}{60--71}.
\newblock


\bibitem[Bao et~al\mbox{.}(2023)]%
        {bao2023tallrec}
\bibfield{author}{\bibinfo{person}{Keqin Bao}, \bibinfo{person}{Jizhi Zhang}, \bibinfo{person}{Yang Zhang}, \bibinfo{person}{Wenjie Wang}, \bibinfo{person}{Fuli Feng}, {and} \bibinfo{person}{Xiangnan He}.} \bibinfo{year}{2023}\natexlab{}.
\newblock \showarticletitle{Tallrec: An effective and efficient tuning framework to align large language model with recommendation}. In \bibinfo{booktitle}{\emph{Proceedings of the 17th ACM Conference on Recommender Systems (RecSys)}}. \bibinfo{pages}{1007--1014}.
\newblock


\bibitem[Beladev et~al\mbox{.}(2016)]%
        {beladev2016recommender}
\bibfield{author}{\bibinfo{person}{Moran Beladev}, \bibinfo{person}{Lior Rokach}, {and} \bibinfo{person}{Bracha Shapira}.} \bibinfo{year}{2016}\natexlab{}.
\newblock \showarticletitle{Recommender systems for product bundling}.
\newblock \bibinfo{journal}{\emph{Knowledge-Based Systems (KBS)}}  \bibinfo{volume}{111} (\bibinfo{year}{2016}), \bibinfo{pages}{193--206}.
\newblock


\bibitem[Cao et~al\mbox{.}(2017)]%
        {cao2017embedding}
\bibfield{author}{\bibinfo{person}{Da Cao}, \bibinfo{person}{Liqiang Nie}, \bibinfo{person}{Xiangnan He}, \bibinfo{person}{Xiaochi Wei}, \bibinfo{person}{Shunzhi Zhu}, {and} \bibinfo{person}{Tat-Seng Chua}.} \bibinfo{year}{2017}\natexlab{}.
\newblock \showarticletitle{Embedding factorization models for jointly recommending items and user generated lists}. In \bibinfo{booktitle}{\emph{Proceedings of the 40th International ACM SIGIR Conference on Research and Development in Information Retrieval (SIGIR)}}. \bibinfo{pages}{585--594}.
\newblock


\bibitem[Chang et~al\mbox{.}(2020)]%
        {chang2020bundle}
\bibfield{author}{\bibinfo{person}{Jianxin Chang}, \bibinfo{person}{Chen Gao}, \bibinfo{person}{Xiangnan He}, \bibinfo{person}{Depeng Jin}, {and} \bibinfo{person}{Yong Li}.} \bibinfo{year}{2020}\natexlab{}.
\newblock \showarticletitle{Bundle recommendation with graph convolutional networks}. In \bibinfo{booktitle}{\emph{Proceedings of the 43rd International ACM SIGIR Conference on Research and Development in Information Retrieval (SIGIR)}}. \bibinfo{pages}{1673--1676}.
\newblock


\bibitem[Chang et~al\mbox{.}(2021)]%
        {chang2021bundle}
\bibfield{author}{\bibinfo{person}{Jianxin Chang}, \bibinfo{person}{Chen Gao}, \bibinfo{person}{Xiangnan He}, \bibinfo{person}{Depeng Jin}, {and} \bibinfo{person}{Yong Li}.} \bibinfo{year}{2021}\natexlab{}.
\newblock \showarticletitle{Bundle recommendation and generation with graph neural networks}.
\newblock \bibinfo{journal}{\emph{IEEE Transactions on Knowledge and Data Engineering (TKDE)}} \bibinfo{volume}{35}, \bibinfo{number}{3} (\bibinfo{year}{2021}), \bibinfo{pages}{2326--2340}.
\newblock


\bibitem[Chen et~al\mbox{.}(2019b)]%
        {chen2019matching}
\bibfield{author}{\bibinfo{person}{Liang Chen}, \bibinfo{person}{Yang Liu}, \bibinfo{person}{Xiangnan He}, \bibinfo{person}{Lianli Gao}, {and} \bibinfo{person}{Zibin Zheng}.} \bibinfo{year}{2019}\natexlab{b}.
\newblock \showarticletitle{Matching user with item set: Collaborative bundle recommendation with deep attention network.}. In \bibinfo{booktitle}{\emph{International Joint Conference on Artificial Intelligence (IJCAI)}}. \bibinfo{pages}{2095--2101}.
\newblock


\bibitem[Chen et~al\mbox{.}(2019a)]%
        {chen2019pog}
\bibfield{author}{\bibinfo{person}{Wen Chen}, \bibinfo{person}{Pipei Huang}, \bibinfo{person}{Jiaming Xu}, \bibinfo{person}{Xin Guo}, \bibinfo{person}{Cheng Guo}, \bibinfo{person}{Fei Sun}, \bibinfo{person}{Chao Li}, \bibinfo{person}{Andreas Pfadler}, \bibinfo{person}{Huan Zhao}, {and} \bibinfo{person}{Binqiang Zhao}.} \bibinfo{year}{2019}\natexlab{a}.
\newblock \showarticletitle{POG: Personalized outfit generation for fashion recommendation at Alibaba iFashion}. In \bibinfo{booktitle}{\emph{Proceedings of the 25th ACM SIGKDD International Conference on Knowledge Discovery and Data Mining (KDD)}}. \bibinfo{pages}{2662--2670}.
\newblock


\bibitem[Cui et~al\mbox{.}(2024)]%
        {cui2024DLLM2Rec}
\bibfield{author}{\bibinfo{person}{Yu Cui}, \bibinfo{person}{Feng Liu}, \bibinfo{person}{Pengbo Wang}, \bibinfo{person}{Bohao Wang}, \bibinfo{person}{Heng Tang}, \bibinfo{person}{Yi Wan}, \bibinfo{person}{Jun Wang}, {and} \bibinfo{person}{Jiawei Chen}.} \bibinfo{year}{2024}\natexlab{}.
\newblock \showarticletitle{Distillation matters: Empowering sequential recommenders to match the performance of large language models}. In \bibinfo{booktitle}{\emph{Proceedings of the 18th ACM Conference on Recommender Systems (RecSys)}}. \bibinfo{pages}{507–517}.
\newblock


\bibitem[Dai et~al\mbox{.}(2023)]%
        {dai2023uncovering}
\bibfield{author}{\bibinfo{person}{Sunhao Dai}, \bibinfo{person}{Ninglu Shao}, \bibinfo{person}{Haiyuan Zhao}, \bibinfo{person}{Weijie Yu}, \bibinfo{person}{Zihua Si}, \bibinfo{person}{Chen Xu}, \bibinfo{person}{Zhongxiang Sun}, \bibinfo{person}{Xiao Zhang}, {and} \bibinfo{person}{Jun Xu}.} \bibinfo{year}{2023}\natexlab{}.
\newblock \showarticletitle{Uncovering chatgpt's capabilities in recommender systems}. In \bibinfo{booktitle}{\emph{Proceedings of the 17th ACM Conference on Recommender Systems (RecSys)}}. \bibinfo{pages}{1126--1132}.
\newblock


\bibitem[Deng et~al\mbox{.}(2021)]%
        {deng2021build}
\bibfield{author}{\bibinfo{person}{Qilin Deng}, \bibinfo{person}{Kai Wang}, \bibinfo{person}{Minghao Zhao}, \bibinfo{person}{Runze Wu}, \bibinfo{person}{Yu Ding}, \bibinfo{person}{Zhene Zou}, \bibinfo{person}{Yue Shang}, \bibinfo{person}{Jianrong Tao}, {and} \bibinfo{person}{Changjie Fan}.} \bibinfo{year}{2021}\natexlab{}.
\newblock \showarticletitle{Build your own bundle - a neural combinatorial optimization method}. In \bibinfo{booktitle}{\emph{Proceedings of the 29th ACM International Conference on Multimedia (MM)}}. \bibinfo{pages}{2625--2633}.
\newblock


\bibitem[Deng et~al\mbox{.}(2020)]%
        {deng2020personalized}
\bibfield{author}{\bibinfo{person}{Qilin Deng}, \bibinfo{person}{Kai Wang}, \bibinfo{person}{Minghao Zhao}, \bibinfo{person}{Zhene Zou}, \bibinfo{person}{Runze Wu}, \bibinfo{person}{Jianrong Tao}, \bibinfo{person}{Changjie Fan}, {and} \bibinfo{person}{Liang Chen}.} \bibinfo{year}{2020}\natexlab{}.
\newblock \showarticletitle{Personalized bundle recommendation in online games}. In \bibinfo{booktitle}{\emph{Proceedings of the 29th ACM International Conference on Information and Knowledge Management (CIKM)}}. \bibinfo{pages}{2381--2388}.
\newblock


\bibitem[Dettmers et~al\mbox{.}(2023)]%
        {dettmers2023qlora}
\bibfield{author}{\bibinfo{person}{Tim Dettmers}, \bibinfo{person}{Artidoro Pagnoni}, \bibinfo{person}{Ari Holtzman}, {and} \bibinfo{person}{Luke Zettlemoyer}.} \bibinfo{year}{2023}\natexlab{}.
\newblock \showarticletitle{QLORA: Efficient finetuning of quantized llms}. In \bibinfo{booktitle}{\emph{Proceedings of the 37th International Conference on Neural Information Processing Systems (NeurIPS)}} \emph{(\bibinfo{series}{NeurIPS '23})}.
\newblock


\bibitem[Dragone et~al\mbox{.}(2018)]%
        {dragone2018no}
\bibfield{author}{\bibinfo{person}{Paolo Dragone}, \bibinfo{person}{Giovanni Pellegrini}, \bibinfo{person}{Michele Vescovi}, \bibinfo{person}{Katya Tentori}, {and} \bibinfo{person}{Andrea Passerini}.} \bibinfo{year}{2018}\natexlab{}.
\newblock \showarticletitle{No more ready-made deals: Constructive recommendation for telco service bundling}. In \bibinfo{booktitle}{\emph{Proceedings of the 12th ACM Conference on Recommender Systems (RecSys)}}. \bibinfo{pages}{163--171}.
\newblock


\bibitem[Fang et~al\mbox{.}(2018)]%
        {fang2018customized}
\bibfield{author}{\bibinfo{person}{Yan Fang}, \bibinfo{person}{Xinyue Xiao}, \bibinfo{person}{Xiaoyu Wang}, {and} \bibinfo{person}{Huiqing Lan}.} \bibinfo{year}{2018}\natexlab{}.
\newblock \showarticletitle{Customized bundle recommendation by association rules of product categories for online supermarkets}. In \bibinfo{booktitle}{\emph{The 3rd International Conference on Data Science in Cyberspace (DSC)}}. \bibinfo{pages}{472--475}.
\newblock


\bibitem[Garfinkel et~al\mbox{.}(2006)]%
        {garfinkel2006design}
\bibfield{author}{\bibinfo{person}{Robert Garfinkel}, \bibinfo{person}{Ram Gopal}, \bibinfo{person}{Arvind Tripathi}, {and} \bibinfo{person}{Fang Yin}.} \bibinfo{year}{2006}\natexlab{}.
\newblock \showarticletitle{Design of a shopbot and recommender system for bundle purchases}.
\newblock \bibinfo{journal}{\emph{Decision Support Systems (DSS)}} \bibinfo{volume}{42}, \bibinfo{number}{3} (\bibinfo{year}{2006}), \bibinfo{pages}{1974--1986}.
\newblock


\bibitem[Ge et~al\mbox{.}(2014)]%
        {ge2014cost}
\bibfield{author}{\bibinfo{person}{Yong Ge}, \bibinfo{person}{Hui Xiong}, \bibinfo{person}{Alexander Tuzhilin}, {and} \bibinfo{person}{Qi Liu}.} \bibinfo{year}{2014}\natexlab{}.
\newblock \showarticletitle{Cost-aware collaborative filtering for travel tour recommendations}.
\newblock \bibinfo{journal}{\emph{ACM Transactions on Information Systems (TOIS)}} \bibinfo{volume}{32}, \bibinfo{number}{1} (\bibinfo{year}{2014}), \bibinfo{pages}{1--31}.
\newblock


\bibitem[Harris and Blair(2006)]%
        {harris2006consumer}
\bibfield{author}{\bibinfo{person}{Judy Harris} {and} \bibinfo{person}{Edward~A Blair}.} \bibinfo{year}{2006}\natexlab{}.
\newblock \showarticletitle{Consumer preference for product bundles: The role of reduced search costs}.
\newblock \bibinfo{journal}{\emph{Journal of the Academy of Marketing Science}} \bibinfo{volume}{34}, \bibinfo{number}{4} (\bibinfo{year}{2006}), \bibinfo{pages}{506--513}.
\newblock


\bibitem[Harte et~al\mbox{.}(2023)]%
        {harte2023leveraging}
\bibfield{author}{\bibinfo{person}{Jesse Harte}, \bibinfo{person}{Wouter Zorgdrager}, \bibinfo{person}{Panos Louridas}, \bibinfo{person}{Asterios Katsifodimos}, \bibinfo{person}{Dietmar Jannach}, {and} \bibinfo{person}{Marios Fragkoulis}.} \bibinfo{year}{2023}\natexlab{}.
\newblock \showarticletitle{Leveraging large language models for sequential recommendation}. In \bibinfo{booktitle}{\emph{Proceedings of the 17th ACM Conference on Recommender Systems (RecSys)}}. \bibinfo{pages}{1096--1102}.
\newblock


\bibitem[He and McAuley(2016)]%
        {he2016ups}
\bibfield{author}{\bibinfo{person}{Ruining He} {and} \bibinfo{person}{Julian McAuley}.} \bibinfo{year}{2016}\natexlab{}.
\newblock \showarticletitle{Ups and downs: Modeling the visual evolution of fashion trends with one-class collaborative filtering}. In \bibinfo{booktitle}{\emph{Proceedings of the 25th International Conference on World Wide Web (WWW)}}. \bibinfo{pages}{507--517}.
\newblock


\bibitem[He et~al\mbox{.}(2019)]%
        {he2019hierarchical}
\bibfield{author}{\bibinfo{person}{Yun He}, \bibinfo{person}{Jianling Wang}, \bibinfo{person}{Wei Niu}, {and} \bibinfo{person}{James Caverlee}.} \bibinfo{year}{2019}\natexlab{}.
\newblock \showarticletitle{A hierarchical self-attentive model for recommending user-generated item lists}. In \bibinfo{booktitle}{\emph{Proceedings of the 28th ACM International Conference on Information and Knowledge Management (CIKM)}}. \bibinfo{pages}{1481--1490}.
\newblock


\bibitem[He et~al\mbox{.}(2020)]%
        {he2020consistency}
\bibfield{author}{\bibinfo{person}{Yun He}, \bibinfo{person}{Yin Zhang}, \bibinfo{person}{Weiwen Liu}, {and} \bibinfo{person}{James Caverlee}.} \bibinfo{year}{2020}\natexlab{}.
\newblock \showarticletitle{Consistency-aware recommendation for user-generated item list continuation}. In \bibinfo{booktitle}{\emph{Proceedings of the 13th International Conference on Web Search and Data Mining (WSDM)}}. \bibinfo{pages}{250--258}.
\newblock


\bibitem[He et~al\mbox{.}(2023)]%
        {he2023large}
\bibfield{author}{\bibinfo{person}{Zhankui He}, \bibinfo{person}{Zhouhang Xie}, \bibinfo{person}{Rahul Jha}, \bibinfo{person}{Harald Steck}, \bibinfo{person}{Dawen Liang}, \bibinfo{person}{Yesu Feng}, \bibinfo{person}{Bodhisattwa Majumder}, \bibinfo{person}{Nathan Kallus}, {and} \bibinfo{person}{Julian Mcauley}.} \bibinfo{year}{2023}\natexlab{}.
\newblock \showarticletitle{Large language models as zero-shot conversational recommenders}. In \bibinfo{booktitle}{\emph{Proceedings of the 32nd ACM International Conference on Information and Knowledge Management (CIKM)}}. \bibinfo{pages}{720–730}.
\newblock


\bibitem[He et~al\mbox{.}(2022)]%
        {he2022bundle}
\bibfield{author}{\bibinfo{person}{Zhankui He}, \bibinfo{person}{Handong Zhao}, \bibinfo{person}{Tong Yu}, \bibinfo{person}{Sungchul Kim}, \bibinfo{person}{Fan Du}, {and} \bibinfo{person}{Julian McAuley}.} \bibinfo{year}{2022}\natexlab{}.
\newblock \showarticletitle{Bundle mcr: Towards conversational bundle recommendation}. In \bibinfo{booktitle}{\emph{Proceedings of the 16th ACM Conference on Recommender Systems (RecSys)}}. \bibinfo{pages}{288--298}.
\newblock


\bibitem[Hinton et~al\mbox{.}(2015)]%
        {hinton2015distilling}
\bibfield{author}{\bibinfo{person}{Geoffrey Hinton}, \bibinfo{person}{Oriol Vinyals}, {and} \bibinfo{person}{Jeff Dean}.} \bibinfo{year}{2015}\natexlab{}.
\newblock \showarticletitle{Distilling the knowledge in a neural network}.
\newblock \bibinfo{journal}{\emph{arXiv preprint arXiv:1503.02531}} (\bibinfo{year}{2015}).
\newblock


\bibitem[Jeon et~al\mbox{.}(2024)]%
        {jeon2024cold}
\bibfield{author}{\bibinfo{person}{Hyunsik Jeon}, \bibinfo{person}{Jong-eun Lee}, \bibinfo{person}{Jeongin Yun}, {and} \bibinfo{person}{U Kang}.} \bibinfo{year}{2024}\natexlab{}.
\newblock \showarticletitle{Cold-start bundle recommendation via popularity-based coalescence and curriculum heating}. In \bibinfo{booktitle}{\emph{Proceedings of the ACM Web Conference 2024}}. \bibinfo{pages}{3277--3286}.
\newblock


\bibitem[Kouki et~al\mbox{.}(2019)]%
        {kouki2019product}
\bibfield{author}{\bibinfo{person}{Pigi Kouki}, \bibinfo{person}{Ilias Fountalis}, \bibinfo{person}{Nikolaos Vasiloglou}, \bibinfo{person}{Nian Yan}, \bibinfo{person}{Unaiza Ahsan}, \bibinfo{person}{Khalifeh~Al Jadda}, {and} \bibinfo{person}{Huiming Qu}.} \bibinfo{year}{2019}\natexlab{}.
\newblock \showarticletitle{Product collection recommendation in online retail}. In \bibinfo{booktitle}{\emph{Proceedings of the 13th ACM Conference on Recommender Systems (RecSys)}}. \bibinfo{pages}{486--490}.
\newblock


\bibitem[Li et~al\mbox{.}(2024)]%
        {li2024quantized}
\bibfield{author}{\bibinfo{person}{Shiyao Li}, \bibinfo{person}{Xuefei Ning}, \bibinfo{person}{Luning Wang}, \bibinfo{person}{Tengxuan Liu}, \bibinfo{person}{Xiangsheng Shi}, \bibinfo{person}{Shengen Yan}, \bibinfo{person}{Guohao Dai}, \bibinfo{person}{Huazhong Yang}, {and} \bibinfo{person}{Yu Wang}.} \bibinfo{year}{2024}\natexlab{}.
\newblock \showarticletitle{Evaluating quantized large language models}. In \bibinfo{booktitle}{\emph{Proceedings of the 41st International Conference on Machine Learning (ICML)}}. Article \bibinfo{articleno}{1144}, \bibinfo{numpages}{45}~pages.
\newblock


\bibitem[Lin et~al\mbox{.}(2024)]%
        {lin2024rella}
\bibfield{author}{\bibinfo{person}{Jianghao Lin}, \bibinfo{person}{Rong Shan}, \bibinfo{person}{Chenxu Zhu}, \bibinfo{person}{Kounianhua Du}, \bibinfo{person}{Bo Chen}, \bibinfo{person}{Shigang Quan}, \bibinfo{person}{Ruiming Tang}, \bibinfo{person}{Yong Yu}, {and} \bibinfo{person}{Weinan Zhang}.} \bibinfo{year}{2024}\natexlab{}.
\newblock \showarticletitle{Rella: Retrieval-enhanced large language models for lifelong sequential behavior comprehension in recommendation}. In \bibinfo{booktitle}{\emph{Proceedings of the ACM Web Conference 2024 (WWW)}}. \bibinfo{pages}{3497--3508}.
\newblock


\bibitem[Liu et~al\mbox{.}(2017)]%
        {liu2017modeling}
\bibfield{author}{\bibinfo{person}{Guannan Liu}, \bibinfo{person}{Yanjie Fu}, \bibinfo{person}{Guoqing Chen}, \bibinfo{person}{Hui Xiong}, {and} \bibinfo{person}{Can Chen}.} \bibinfo{year}{2017}\natexlab{}.
\newblock \showarticletitle{Modeling buying motives for personalized product bundle recommendation}.
\newblock \bibinfo{journal}{\emph{ACM Transactions on Knowledge Discovery from Data (TKDD)}} \bibinfo{volume}{11}, \bibinfo{number}{3} (\bibinfo{year}{2017}), \bibinfo{pages}{1--26}.
\newblock


\bibitem[Liu et~al\mbox{.}(2011)]%
        {liu2011personalized}
\bibfield{author}{\bibinfo{person}{Qi Liu}, \bibinfo{person}{Yong Ge}, \bibinfo{person}{Zhongmou Li}, \bibinfo{person}{Enhong Chen}, {and} \bibinfo{person}{Hui Xiong}.} \bibinfo{year}{2011}\natexlab{}.
\newblock \showarticletitle{Personalized travel package recommendation}. In \bibinfo{booktitle}{\emph{IEEE 11th International Conference on Data Mining (ICDM)}}. \bibinfo{pages}{407--416}.
\newblock


\bibitem[Liu et~al\mbox{.}(2025)]%
        {liu2025harnessing}
\bibfield{author}{\bibinfo{person}{Xiaohao Liu}, \bibinfo{person}{Jie Wu}, \bibinfo{person}{Zhulin Tao}, \bibinfo{person}{Yunshan Ma}, \bibinfo{person}{Yinwei Wei}, {and} \bibinfo{person}{Tat-seng Chua}.} \bibinfo{year}{2025}\natexlab{}.
\newblock \showarticletitle{Fine-tuning multimodal large language models for product bundling}. In \bibinfo{booktitle}{\emph{Proceedings of the 31st ACM SIGKDD Conference on Knowledge Discovery and Data Mining (KDD)}}. \bibinfo{pages}{848–858}.
\newblock


\bibitem[Liu et~al\mbox{.}(2014)]%
        {liu2014recommending}
\bibfield{author}{\bibinfo{person}{Yidan Liu}, \bibinfo{person}{Min Xie}, {and} \bibinfo{person}{Laks~VS Lakshmanan}.} \bibinfo{year}{2014}\natexlab{}.
\newblock \showarticletitle{Recommending user generated item lists}. In \bibinfo{booktitle}{\emph{Proceedings of the 8th ACM Conference on Recommender Systems (RecSys)}}. \bibinfo{pages}{185--192}.
\newblock


\bibitem[Ma et~al\mbox{.}(2024a)]%
        {ma2024multicbr}
\bibfield{author}{\bibinfo{person}{Yunshan Ma}, \bibinfo{person}{Yingzhi He}, \bibinfo{person}{Xiang Wang}, \bibinfo{person}{Yinwei Wei}, \bibinfo{person}{Xiaoyu Du}, \bibinfo{person}{Yuyangzi Fu}, {and} \bibinfo{person}{Tat-Seng Chua}.} \bibinfo{year}{2024}\natexlab{a}.
\newblock \showarticletitle{Multicbr: Multi-view contrastive learning for bundle recommendation}.
\newblock \bibinfo{journal}{\emph{ACM Transactions on Information Systems (TOIS)}} \bibinfo{volume}{42}, \bibinfo{number}{4} (\bibinfo{year}{2024}), \bibinfo{pages}{1--23}.
\newblock


\bibitem[Ma et~al\mbox{.}(2022)]%
        {ma2022crosscbr}
\bibfield{author}{\bibinfo{person}{Yunshan Ma}, \bibinfo{person}{Yingzhi He}, \bibinfo{person}{An Zhang}, \bibinfo{person}{Xiang Wang}, {and} \bibinfo{person}{Tat-Seng Chua}.} \bibinfo{year}{2022}\natexlab{}.
\newblock \showarticletitle{CrossCBR: Cross-view contrastive learning for bundle recommendation}. In \bibinfo{booktitle}{\emph{Proceedings of the 28th ACM SIGKDD Conference on Knowledge Discovery and Data Mining (KDD)}}. \bibinfo{pages}{1233--1241}.
\newblock


\bibitem[Ma et~al\mbox{.}(2024b)]%
        {ma2024cirp}
\bibfield{author}{\bibinfo{person}{Yunshan Ma}, \bibinfo{person}{Yingzhi He}, \bibinfo{person}{Wenjun Zhong}, \bibinfo{person}{Xiang Wang}, \bibinfo{person}{Roger Zimmermann}, {and} \bibinfo{person}{Tat-Seng Chua}.} \bibinfo{year}{2024}\natexlab{b}.
\newblock \showarticletitle{CIRP: Cross-item relational pre-training for multimodal product bundling}. In \bibinfo{booktitle}{\emph{Proceedings of the 32nd ACM International Conference on Multimedia (MM)}}. \bibinfo{pages}{9641--9649}.
\newblock


\bibitem[Ma et~al\mbox{.}(2024c)]%
        {ma2024leveraging}
\bibfield{author}{\bibinfo{person}{Yunshan Ma}, \bibinfo{person}{Xiaohao Liu}, \bibinfo{person}{Yinwei Wei}, \bibinfo{person}{Zhulin Tao}, \bibinfo{person}{Xiang Wang}, {and} \bibinfo{person}{Tat-Seng Chua}.} \bibinfo{year}{2024}\natexlab{c}.
\newblock \showarticletitle{Leveraging multimodal features and item-level user feedback for bundle construction}. In \bibinfo{booktitle}{\emph{Proceedings of the 17th ACM International Conference on Web Search and Data Mining (WSDM)}}. \bibinfo{pages}{510--519}.
\newblock


\bibitem[Madaan et~al\mbox{.}(2023)]%
        {madaan2023self}
\bibfield{author}{\bibinfo{person}{Aman Madaan}, \bibinfo{person}{Niket Tandon}, \bibinfo{person}{Prakhar Gupta}, \bibinfo{person}{Skyler Hallinan}, \bibinfo{person}{Luyu Gao}, \bibinfo{person}{Sarah Wiegreffe}, \bibinfo{person}{Uri Alon}, \bibinfo{person}{Nouha Dziri}, \bibinfo{person}{Shrimai Prabhumoye}, \bibinfo{person}{Yiming Yang}, {et~al\mbox{.}}} \bibinfo{year}{2023}\natexlab{}.
\newblock \showarticletitle{Self-refine: Iterative refinement with self-feedback}. In \bibinfo{booktitle}{\emph{Advances in Neural Information Processing Systems (NeurIPS)}}.
\newblock


\bibitem[Pathak et~al\mbox{.}(2017)]%
        {pathak2017generating}
\bibfield{author}{\bibinfo{person}{Apurva Pathak}, \bibinfo{person}{Kshitiz Gupta}, {and} \bibinfo{person}{Julian McAuley}.} \bibinfo{year}{2017}\natexlab{}.
\newblock \showarticletitle{Generating and personalizing bundle recommendations on steam}. In \bibinfo{booktitle}{\emph{Proceedings of the 40th International ACM SIGIR Conference on Research and Development in Information Retrieval (SIGIR)}}. \bibinfo{pages}{1073--1076}.
\newblock


\bibitem[Ren et~al\mbox{.}(2023)]%
        {ren2023distillation}
\bibfield{author}{\bibinfo{person}{Yuyang Ren}, \bibinfo{person}{Zhang Haonan}, \bibinfo{person}{Luoyi Fu}, \bibinfo{person}{Xinbing Wang}, {and} \bibinfo{person}{Chenghu Zhou}.} \bibinfo{year}{2023}\natexlab{}.
\newblock \showarticletitle{Distillation-enhanced graph masked autoencoders for bundle recommendation}. In \bibinfo{booktitle}{\emph{Proceedings of the 46th International ACM SIGIR Conference on Research and Development in Information Retrieval (SIGIR)}}. \bibinfo{pages}{1660--1669}.
\newblock


\bibitem[Rendle et~al\mbox{.}(2009)]%
        {rendle2012bpr}
\bibfield{author}{\bibinfo{person}{Steffen Rendle}, \bibinfo{person}{Christoph Freudenthaler}, \bibinfo{person}{Zeno Gantner}, {and} \bibinfo{person}{Lars Schmidt-Thieme}.} \bibinfo{year}{2009}\natexlab{}.
\newblock \showarticletitle{BPR: Bayesian personalized ranking from implicit feedback}. In \bibinfo{booktitle}{\emph{Proceedings of the Twenty-Fifth Conference on Uncertainty in Artificial Intelligence (UAI)}}. \bibinfo{pages}{452--461}.
\newblock


\bibitem[Salemi et~al\mbox{.}(2024)]%
        {salemi2024lamp}
\bibfield{author}{\bibinfo{person}{Alireza Salemi}, \bibinfo{person}{Surya Kallumadi}, {and} \bibinfo{person}{Hamed Zamani}.} \bibinfo{year}{2024}\natexlab{}.
\newblock \showarticletitle{Optimization methods for personalizing large language models through retrieval augmentation}. In \bibinfo{booktitle}{\emph{Proceedings of the 47th International ACM SIGIR Conference on Research and Development in Information Retrieval (SIGIR)}}. \bibinfo{pages}{752--762}.
\newblock


\bibitem[Sar~Shalom et~al\mbox{.}(2016)]%
        {sar2016beyond}
\bibfield{author}{\bibinfo{person}{Oren Sar~Shalom}, \bibinfo{person}{Noam Koenigstein}, \bibinfo{person}{Ulrich Paquet}, {and} \bibinfo{person}{Hastagiri~P Vanchinathan}.} \bibinfo{year}{2016}\natexlab{}.
\newblock \showarticletitle{Beyond collaborative filtering: The list recommendation problem}. In \bibinfo{booktitle}{\emph{Proceedings of the 25th International Conference on World Wide Web (WWW))}}. \bibinfo{pages}{63--72}.
\newblock


\bibitem[Sun et~al\mbox{.}(2024d)]%
        {wen2024prmkd}
\bibfield{author}{\bibinfo{person}{Wenqi Sun}, \bibinfo{person}{Ruobing Xie}, \bibinfo{person}{Junjie Zhang}, \bibinfo{person}{Wayne~Xin Zhao}, \bibinfo{person}{Leyu Lin}, {and} \bibinfo{person}{Ji-Rong Wen}.} \bibinfo{year}{2024}\natexlab{d}.
\newblock \showarticletitle{Distillation is all you need for practically using different pre-trained recommendation models}.
\newblock \bibinfo{journal}{\emph{CoRR}} (\bibinfo{year}{2024}).
\newblock


\bibitem[Sun et~al\mbox{.}(2024a)]%
        {sun2024revisiting}
\bibfield{author}{\bibinfo{person}{Zhu Sun}, \bibinfo{person}{Kaidong Feng}, \bibinfo{person}{Jie Yang}, \bibinfo{person}{Hui Fang}, \bibinfo{person}{Xinghua Qu}, \bibinfo{person}{Yew-Soon Ong}, {and} \bibinfo{person}{Wenyuan Liu}.} \bibinfo{year}{2024}\natexlab{a}.
\newblock \showarticletitle{Revisiting bundle recommendation for intent-aware product bundling}.
\newblock \bibinfo{journal}{\emph{ACM Transactions on Recommender Systems (TORS)}} (\bibinfo{year}{2024}).
\newblock


\bibitem[Sun et~al\mbox{.}(2024b)]%
        {sun2024adaptive}
\bibfield{author}{\bibinfo{person}{Zhu Sun}, \bibinfo{person}{Kaidong Feng}, \bibinfo{person}{Jie Yang}, \bibinfo{person}{Xinghua Qu}, \bibinfo{person}{Hui Fang}, \bibinfo{person}{Yew-Soon Ong}, {and} \bibinfo{person}{Wenyuan Liu}.} \bibinfo{year}{2024}\natexlab{b}.
\newblock \showarticletitle{Adaptive in-context learning with large language models for bundle generation}. In \bibinfo{booktitle}{\emph{Proceedings of the 47th International ACM SIGIR Conference on Research and Development in Information Retrieval (SIGIR)}}. \bibinfo{pages}{966--976}.
\newblock


\bibitem[Sun et~al\mbox{.}(2024c)]%
        {sun2024large}
\bibfield{author}{\bibinfo{person}{Zhu Sun}, \bibinfo{person}{Hongyang Liu}, \bibinfo{person}{Xinghua Qu}, \bibinfo{person}{Kaidong Feng}, \bibinfo{person}{Yan Wang}, {and} \bibinfo{person}{Yew~Soon Ong}.} \bibinfo{year}{2024}\natexlab{c}.
\newblock \showarticletitle{Large language models for intent-driven session recommendations}. In \bibinfo{booktitle}{\emph{Proceedings of the 47th International ACM SIGIR Conference on Research and Development in Information Retrieval (SIGIR)}}. \bibinfo{pages}{324--334}.
\newblock


\bibitem[Sun et~al\mbox{.}(2022)]%
        {sun2022revisiting}
\bibfield{author}{\bibinfo{person}{Zhu Sun}, \bibinfo{person}{Jie Yang}, \bibinfo{person}{Kaidong Feng}, \bibinfo{person}{Hui Fang}, \bibinfo{person}{Xinghua Qu}, {and} \bibinfo{person}{Yew~Soon Ong}.} \bibinfo{year}{2022}\natexlab{}.
\newblock \showarticletitle{Revisiting bundle recommendation: Datasets, tasks, challenges and opportunities for intent-aware product bundling}. In \bibinfo{booktitle}{\emph{Proceedings of the 45th International ACM SIGIR Conference on Research and Development in Information Retrieval (SIGIR)}}. \bibinfo{pages}{2900--2911}.
\newblock


\bibitem[Wang et~al\mbox{.}(2023b)]%
        {wang2023generative}
\bibfield{author}{\bibinfo{person}{Wenjie Wang}, \bibinfo{person}{Xinyu Lin}, \bibinfo{person}{Fuli Feng}, \bibinfo{person}{Xiangnan He}, {and} \bibinfo{person}{Tat-Seng Chua}.} \bibinfo{year}{2023}\natexlab{b}.
\newblock \showarticletitle{Generative recommendation: Towards next-generation recommender paradigm}.
\newblock \bibinfo{journal}{\emph{arXiv preprint arXiv:2304.03516}} (\bibinfo{year}{2023}).
\newblock


\bibitem[Wang et~al\mbox{.}(2024a)]%
        {wang2024rdrec}
\bibfield{author}{\bibinfo{person}{Xinfeng Wang}, \bibinfo{person}{Jin Cui}, \bibinfo{person}{Yoshimi Suzuki}, {and} \bibinfo{person}{Fumiyo Fukumoto}.} \bibinfo{year}{2024}\natexlab{a}.
\newblock \showarticletitle{{RDR}ec: Rationale distillation for {LLM}-based recommendation}. In \bibinfo{booktitle}{\emph{Proceedings of the 62nd Annual Meeting of the Association for Computational Linguistics (ACL)}}. \bibinfo{pages}{65--74}.
\newblock


\bibitem[Wang et~al\mbox{.}(2023c)]%
        {wang2023self}
\bibfield{author}{\bibinfo{person}{Xuezhi Wang}, \bibinfo{person}{Jason Wei}, \bibinfo{person}{Dale Schuurmans}, \bibinfo{person}{Quoc Le}, \bibinfo{person}{Ed Chi}, \bibinfo{person}{Sharan Narang}, \bibinfo{person}{Aakanksha Chowdhery}, {and} \bibinfo{person}{Denny Zhou}.} \bibinfo{year}{2023}\natexlab{c}.
\newblock \showarticletitle{Self-consistency improves chain of thought reasoning in language models}. In \bibinfo{booktitle}{\emph{Proceedings of the 11th International Conference on Learning Representations (ICLR)}}.
\newblock


\bibitem[Wang et~al\mbox{.}(2023a)]%
        {wang2023enhancing}
\bibfield{author}{\bibinfo{person}{Yan Wang}, \bibinfo{person}{Zhixuan Chu}, \bibinfo{person}{Xin Ouyang}, \bibinfo{person}{Simeng Wang}, \bibinfo{person}{Hongyan Hao}, \bibinfo{person}{Yue Shen}, \bibinfo{person}{Jinjie Gu}, \bibinfo{person}{Siqiao Xue}, \bibinfo{person}{James~Y Zhang}, \bibinfo{person}{Qing Cui}, {et~al\mbox{.}}} \bibinfo{year}{2023}\natexlab{a}.
\newblock \showarticletitle{Enhancing recommender systems with large language model reasoning graphs}.
\newblock \bibinfo{journal}{\emph{arXiv preprint arXiv:2308.10835}} (\bibinfo{year}{2023}).
\newblock


\bibitem[Wang et~al\mbox{.}(2024b)]%
        {wang2024SLIM}
\bibfield{author}{\bibinfo{person}{Yuling Wang}, \bibinfo{person}{Changxin Tian}, \bibinfo{person}{Binbin Hu}, \bibinfo{person}{Yanhua Yu}, \bibinfo{person}{Ziqi Liu}, \bibinfo{person}{Zhiqiang Zhang}, \bibinfo{person}{Jun Zhou}, \bibinfo{person}{Liang Pang}, {and} \bibinfo{person}{Xiao Wang}.} \bibinfo{year}{2024}\natexlab{b}.
\newblock \showarticletitle{Can small language models be good reasoners for sequential recommendation?}. In \bibinfo{booktitle}{\emph{Proceedings of the ACM Web Conference 2024 (WWW)}}.
\newblock


\bibitem[Wei et~al\mbox{.}(2022)]%
        {wei2022towards}
\bibfield{author}{\bibinfo{person}{Penghui Wei}, \bibinfo{person}{Shaoguo Liu}, \bibinfo{person}{Xuanhua Yang}, \bibinfo{person}{Liang Wang}, {and} \bibinfo{person}{Bo Zheng}.} \bibinfo{year}{2022}\natexlab{}.
\newblock \showarticletitle{Towards personalized bundle creative generation with contrastive non-autoregressive decoding}. In \bibinfo{booktitle}{\emph{Proceedings of the 45th International ACM SIGIR Conference on Research and Development in Information Retrieval (SIGIR)}}. \bibinfo{pages}{2634--2638}.
\newblock


\bibitem[Wei et~al\mbox{.}(2023)]%
        {wei2023strategy}
\bibfield{author}{\bibinfo{person}{Yinwei Wei}, \bibinfo{person}{Xiaohao Liu}, \bibinfo{person}{Yunshan Ma}, \bibinfo{person}{Xiang Wang}, \bibinfo{person}{Liqiang Nie}, {and} \bibinfo{person}{Tat-Seng Chua}.} \bibinfo{year}{2023}\natexlab{}.
\newblock \showarticletitle{Strategy-aware bundle recommender system}. In \bibinfo{booktitle}{\emph{Proceedings of the 46th International ACM SIGIR Conference on Research and Development in Information Retrieval (SIGIR)}}. \bibinfo{pages}{1198--1207}.
\newblock


\bibitem[Wu et~al\mbox{.}(2021)]%
        {wu2021newsbert}
\bibfield{author}{\bibinfo{person}{Chuhan Wu}, \bibinfo{person}{Fangzhao Wu}, \bibinfo{person}{Yang Yu}, \bibinfo{person}{Tao Qi}, \bibinfo{person}{Yongfeng Huang}, {and} \bibinfo{person}{Qi Liu}.} \bibinfo{year}{2021}\natexlab{}.
\newblock \showarticletitle{{N}ews{BERT}: Distilling pre-trained language model for intelligent news application}. In \bibinfo{booktitle}{\emph{Findings of the Association for Computational Linguistics: EMNLP 2021}}.
\newblock


\bibitem[Wu et~al\mbox{.}(2024)]%
        {wu2024exploring}
\bibfield{author}{\bibinfo{person}{Likang Wu}, \bibinfo{person}{Zhaopeng Qiu}, \bibinfo{person}{Zhi Zheng}, \bibinfo{person}{Hengshu Zhu}, {and} \bibinfo{person}{Enhong Chen}.} \bibinfo{year}{2024}\natexlab{}.
\newblock \showarticletitle{Exploring large language model for graph data understanding in online job recommendations}. In \bibinfo{booktitle}{\emph{Proceedings of the AAAI Conference on Artificial Intelligence (AAAI)}}, Vol.~\bibinfo{volume}{38}. \bibinfo{pages}{9178--9186}.
\newblock


\bibitem[Wu et~al\mbox{.}(2023)]%
        {wu2023survey}
\bibfield{author}{\bibinfo{person}{Likang Wu}, \bibinfo{person}{Zhi Zheng}, \bibinfo{person}{Zhaopeng Qiu}, \bibinfo{person}{Hao Wang}, \bibinfo{person}{Hongchao Gu}, \bibinfo{person}{Tingjia Shen}, \bibinfo{person}{Chuan Qin}, \bibinfo{person}{Chen Zhu}, \bibinfo{person}{Hengshu Zhu}, \bibinfo{person}{Qi Liu}, {et~al\mbox{.}}} \bibinfo{year}{2023}\natexlab{}.
\newblock \showarticletitle{A survey on large language models for recommendation}.
\newblock \bibinfo{journal}{\emph{arXiv preprint arXiv:2305.19860}} (\bibinfo{year}{2023}).
\newblock


\bibitem[Xie et~al\mbox{.}(2010)]%
        {xie2010breaking}
\bibfield{author}{\bibinfo{person}{Min Xie}, \bibinfo{person}{Laks~VS Lakshmanan}, {and} \bibinfo{person}{Peter~T Wood}.} \bibinfo{year}{2010}\natexlab{}.
\newblock \showarticletitle{Breaking out of the box of recommendations: From items to packages}. In \bibinfo{booktitle}{\emph{Proceedings of the 4th ACM Conference on Recommender Systems (RecSys)}}. \bibinfo{pages}{151--158}.
\newblock


\bibitem[Xie et~al\mbox{.}(2014)]%
        {xie2014generating}
\bibfield{author}{\bibinfo{person}{Min Xie}, \bibinfo{person}{Laks~VS Lakshmanan}, {and} \bibinfo{person}{Peter~T Wood}.} \bibinfo{year}{2014}\natexlab{}.
\newblock \showarticletitle{Generating top-k packages via preference elicitation}.
\newblock \bibinfo{journal}{\emph{Proceedings of the VLDB Endowment (VLDB)}} \bibinfo{volume}{7}, \bibinfo{number}{14} (\bibinfo{year}{2014}), \bibinfo{pages}{1941--1952}.
\newblock


\bibitem[Xu et~al\mbox{.}(2024)]%
        {xu2024diffusion}
\bibfield{author}{\bibinfo{person}{Yiyan Xu}, \bibinfo{person}{Wenjie Wang}, \bibinfo{person}{Fuli Feng}, \bibinfo{person}{Yunshan Ma}, \bibinfo{person}{Jizhi Zhang}, {and} \bibinfo{person}{Xiangnan He}.} \bibinfo{year}{2024}\natexlab{}.
\newblock \showarticletitle{Diffusion models for generative outfit recommendation}. In \bibinfo{booktitle}{\emph{Proceedings of the 47th international ACM SIGIR conference on research and development in information retrieval (SIGIR)}}. \bibinfo{pages}{1350--1359}.
\newblock


\bibitem[Yang et~al\mbox{.}(2012)]%
        {yang2012bundle}
\bibfield{author}{\bibinfo{person}{De-Nian Yang}, \bibinfo{person}{Wang-Chien Lee}, \bibinfo{person}{Nai-Hui Chia}, \bibinfo{person}{Mao Ye}, {and} \bibinfo{person}{Hui-Ju Hung}.} \bibinfo{year}{2012}\natexlab{}.
\newblock \showarticletitle{On bundle configuration for viral marketing in social networks}. In \bibinfo{booktitle}{\emph{Proceedings of the 21st ACM International Conference on Information and Knowledge Management (CIKM)}}. \bibinfo{pages}{2234--2238}.
\newblock


\bibitem[Yu et~al\mbox{.}(2021)]%
        {yu2021tiny}
\bibfield{author}{\bibinfo{person}{Yang Yu}, \bibinfo{person}{Fangzhao Wu}, \bibinfo{person}{Chuhan Wu}, \bibinfo{person}{Jingwei Yi}, {and} \bibinfo{person}{Qi Liu}.} \bibinfo{year}{2021}\natexlab{}.
\newblock \showarticletitle{Tiny-newsrec: Effective and efficient plm-based news recommendation}.
\newblock \bibinfo{journal}{\emph{arXiv preprint arXiv:2112.00944}} (\bibinfo{year}{2021}).
\newblock


\bibitem[Yuan et~al\mbox{.}(2021)]%
        {yuan2021improving}
\bibfield{author}{\bibinfo{person}{Xu Yuan}, \bibinfo{person}{Hongshen Chen}, \bibinfo{person}{Yonghao Song}, \bibinfo{person}{Xiaofang Zhao}, \bibinfo{person}{Zhuoye Ding}, \bibinfo{person}{Zhen He}, {and} \bibinfo{person}{Bo Long}.} \bibinfo{year}{2021}\natexlab{}.
\newblock \showarticletitle{Improving sequential recommendation consistency with self-supervised imitation}.
\newblock \bibinfo{journal}{\emph{arXiv preprint arXiv:2106.14031}} (\bibinfo{year}{2021}).
\newblock


\bibitem[Zhai et~al\mbox{.}(2023)]%
        {zhai2023knowledge}
\bibfield{author}{\bibinfo{person}{Jianyang Zhai}, \bibinfo{person}{Xiawu Zheng}, \bibinfo{person}{Chang-Dong Wang}, \bibinfo{person}{Hui Li}, {and} \bibinfo{person}{Yonghong Tian}.} \bibinfo{year}{2023}\natexlab{}.
\newblock \showarticletitle{Knowledge prompt-tuning for sequential recommendation}. In \bibinfo{booktitle}{\emph{Proceedings of the 29th ACM International Conference on Multimedia (MM)}}. \bibinfo{pages}{6451–6461}.
\newblock


\bibitem[Zhang et~al\mbox{.}(2023)]%
        {zhang2023chatgpt}
\bibfield{author}{\bibinfo{person}{Jizhi Zhang}, \bibinfo{person}{Keqin Bao}, \bibinfo{person}{Yang Zhang}, \bibinfo{person}{Wenjie Wang}, \bibinfo{person}{Fuli Feng}, {and} \bibinfo{person}{Xiangnan He}.} \bibinfo{year}{2023}\natexlab{}.
\newblock \showarticletitle{Is chatgpt fair for recommendation? Evaluating fairness in large language model recommendation}. In \bibinfo{booktitle}{\emph{Proceedings of the 17th ACM Conference on Recommender Systems (RecSys)}}. \bibinfo{pages}{993--999}.
\newblock


\bibitem[Zhao et~al\mbox{.}(2022)]%
        {zhao2022multi}
\bibfield{author}{\bibinfo{person}{Sen Zhao}, \bibinfo{person}{Wei Wei}, \bibinfo{person}{Ding Zou}, {and} \bibinfo{person}{Xianling Mao}.} \bibinfo{year}{2022}\natexlab{}.
\newblock \showarticletitle{Multi-view intent disentangle graph networks for bundle recommendation}. In \bibinfo{booktitle}{\emph{Proceedings of the AAAI Conference on Artificial Intelligence (AAAI)}}. \bibinfo{pages}{4379--4387}.
\newblock


\bibitem[Zhu et~al\mbox{.}(2014)]%
        {zhu2014bundle}
\bibfield{author}{\bibinfo{person}{Tao Zhu}, \bibinfo{person}{Patrick Harrington}, \bibinfo{person}{Junjun Li}, {and} \bibinfo{person}{Lei Tang}.} \bibinfo{year}{2014}\natexlab{}.
\newblock \showarticletitle{Bundle recommendation in ecommerce}. In \bibinfo{booktitle}{\emph{Proceedings of the 37th International ACM SIGIR Conference on Research and Development in Information Retrieval (SIGIR)}}. \bibinfo{pages}{657--666}.
\newblock


\bibitem[Zou et~al\mbox{.}(2023)]%
        {zou2023towards}
\bibfield{author}{\bibinfo{person}{Ding Zou}, \bibinfo{person}{Sen Zhao}, \bibinfo{person}{Wei Wei}, \bibinfo{person}{Xian-ling Mao}, \bibinfo{person}{Ruixuan Li}, \bibinfo{person}{Dangyang Chen}, \bibinfo{person}{Rui Fang}, {and} \bibinfo{person}{Yuanyuan Fu}.} \bibinfo{year}{2023}\natexlab{}.
\newblock \showarticletitle{Towards hierarchical intent disentanglement for bundle recommendation}.
\newblock \bibinfo{journal}{\emph{IEEE Transactions on Knowledge and Data Engineering (TKDE)}} \bibinfo{volume}{36}, \bibinfo{number}{7} (\bibinfo{year}{2023}), \bibinfo{pages}{3556--3567}.
\newblock


\end{thebibliography}










\end{document}